\documentclass[journal]{IEEEtran}

\usepackage{amsfonts}
\usepackage{color}
\usepackage{bm}
\usepackage{mathrsfs}
\usepackage{multirow}
\usepackage{balance}
\usepackage{url}
\newcommand{\sset}[1]{\bm{\mathcal{\MakeUppercase{#1}}}}

\ifCLASSINFOpdf
  \usepackage[pdftex]{graphicx}
  \graphicspath{{pdf/}{jpeg/}}
  \DeclareGraphicsExtensions{.pdf,.jpeg,.png}
\else
\fi
\usepackage{amsmath}
\usepackage{algorithm}
\usepackage{algorithmic}

\ifCLASSOPTIONcompsoc
 \usepackage[caption=false,font=normalsize,labelfont=sf,textfont=sf]{subfig}
\else
 \usepackage[caption=false,font=footnotesize]{subfig}
\fi

\usepackage{stfloats}
\begin{document}

%
\title{Benchmark data and method for real-time people counting in cluttered scenes using depth sensors}
%
%
%

\author{ShiJie~Sun,
		Naveed~Akhtar, 
		HuanSheng~Song,
        ChaoYang~Zhang,
        JianXin~Li,
        Ajmal~Mian
 \thanks{This work was supported by ARC Grant  DP160101458, the National Natural Science Foundation of China (Grant No. 61572083), the Joint Fund of Ministry of Education of China (Grant No. 6141A02022610), the Fundamental Research Funds for the Central Universities (Grant No. 310824171003) and China Scholarship Council (CSC).}
 \thanks{ShiJie~Sun, HuanSheng~Song, ChaoYang~Zhang are with the School
of Information Engineering, Chang'an University, Xi'an 710000, ShannXi, China. e-mail: (shijieSun@chd.edu.cn, hshsong@chd.edu.cn, zhaoyang\_zh@chd.edu.cn).}
\thanks{Naveed~Akhtar, JianXin~Li, Ajmal~Mian are with the School of Computer Science and Software Engineering, The University of Western Australia, Crawley, WA, 6009, Australia. e-mail: (naveed.akhtar@uwa.edu.au, jianxin.li@uwa.edu.au, ajmal.mian@uwa.edu.au).}
}

%
%

\markboth{Journal of \LaTeX\ Class Files,~Vol.~14, No.~8, April~2018}%
{Shell \MakeLowercase{\textit{et al.}}: Bare Demo of IEEEtran.cls for IEEE Journals}
%



\maketitle

\begin{abstract}
Vision-based automatic counting of people has widespread applications in intelligent transportation systems, security, and logistics. 
However, there is currently no large-scale public dataset for benchmarking approaches on this problem.  
This work fills this gap by introducing the first real-world RGB-D \textbf{P}eople \textbf{C}ounting \textbf{D}ata\textbf{S}et (PCDS) containing over $4,500$ videos recorded at the entrance doors of buses in normal and cluttered conditions. 
It also proposes an efficient method for counting people in real-world cluttered scenes related to public transportations using depth videos.
The proposed method computes a point cloud from the depth video frame and re-projects it onto the ground plane to normalize the depth information. The resulting depth image is analyzed for identifying potential human heads.
The human head proposals are meticulously refined using a 3D human model. The proposals in each frame of the continuous video stream are tracked to trace their trajectories. The trajectories are again refined to ascertain reliable counting.
People are eventually counted by accumulating the head trajectories leaving the scene. To enable effective head and trajectory identification, we also propose two different compound features. A thorough evaluation on PCDS demonstrates that our technique is able to count people in cluttered scenes with high accuracy at 45 fps on a 1.7 GHz processor, and hence it can be deployed for effective real-time people counting for intelligent transportation systems.  

\end{abstract}


%
\IEEEpeerreviewmaketitle

\vspace{-3mm}
\section{Introduction}
Automatic people counting in real-time has multiple applications in intelligent public transportation systems~\cite{6817563}, \cite{7837703}, \cite{ceder1984bus}. On effective method to reliably accomplish this task is to directly analyze continuous video streams of vehicle entrance and exit doors, and perform automatic counting of people in those videos.  
For intelligent public transportation systems, such as buses with on-line monitoring, knowing the number of people entering and leaving the transport can be used in e.g.~dynamic planning to avoid congestion. It also promises significant economic benefits by improving transportation scheduling in accordance with human traffic on stations at different hours of operation. 

Computer Vision techniques are well-suited to the problem of automatic people counting for public transportations. However, using conventional RGB videos for this purpose is challenged by multiple issues resulting from real-world conditions such as clutter, occlusions, illumination variations, handling shadows etc.
In comparison to the conventional video systems, RGB-D cameras (e.g.~Kinect V1~\cite{Zhang2012c}, Prime Sense Camera~\cite{Freedman2010}) can mitigate these issues by providing `depth' information of the scene in addition to its color video.
Nevertheless, effective people counting in real-world conditions using depth information still remains a largely unsolved problem due to noise and occlusion~\cite{Mallick2014}.

Vision-based people counting is a comprehensive task that involves object detection, recognition, and tracking. Existing approaches in this area can be broadly categorized into three classes: (a)~regression-based methods, e.g.~\cite{Barandiaran2008a}, \cite{Fradi2012} (b)~clustering-based methods, e.g.~\cite{Antonini2006}, \cite{Topkaya2014}, and (c)~detection-based methods, e.g.~\cite{Zeng2010}, \cite{DelPizzo2016a}. Regression-based methods aim at learning a regression function using features of detection regions and exploit that for counting. Clustering-based methods track a set of features of target objects, and cluster their trajectories for counting them. Detection-based methods have a common pipeline, comprising foreground extraction, target localization, tracking, and trajectory classification. We can further divide these methods based on the data types they use e.g.~color/depth/hybrid video methods (see Section~\ref{sec:RW} for the details).
Although useful, the above mentioned approaches face some common problems while counting people under  practical conditions in real-time, which include; restriction of camera angles~\cite{Chen2012a, Li2017, Gao2016}; 
computational inefficiency \cite{Liu2017}, and failing to handle cluttered scenes~\cite{Kocak2017}. 
Moreover, to the best of our knowledge, there is currently no large-scale public dataset available to benchmark methods for real-world people counting.

In this paper, we first introduce a large-scale dataset for the problem of counting people in real-world scenes of bus entrance/exit doors.
The dataset, called \textbf{P}eople \textbf{C}ounting \textbf{D}ata\textbf{S}et (PCDS) contains $4,689$ videos acquired with the Kinect V1 camera~\cite{Zhang2012c} that contains RGB and depth sensors. The dataset can be publicly downloaded using the following URL (https://github.com/shijieS/people-counting-dataset.git). Each video in the dataset is labeled with the number of people entering or exiting the bus door.
The data has been collected on three different bus routes at different times of the day on 4 - 6 different days, and presents large variations in terms of illumination, occlusion, clutter and noise. 
As a second major contribution, we propose a real-time method for counting people passing through cluttered scenes. The proposed method uses the depth video stream for people counting.  

\begin{figure*}[tb]
\centering
\includegraphics[width=6.3in]{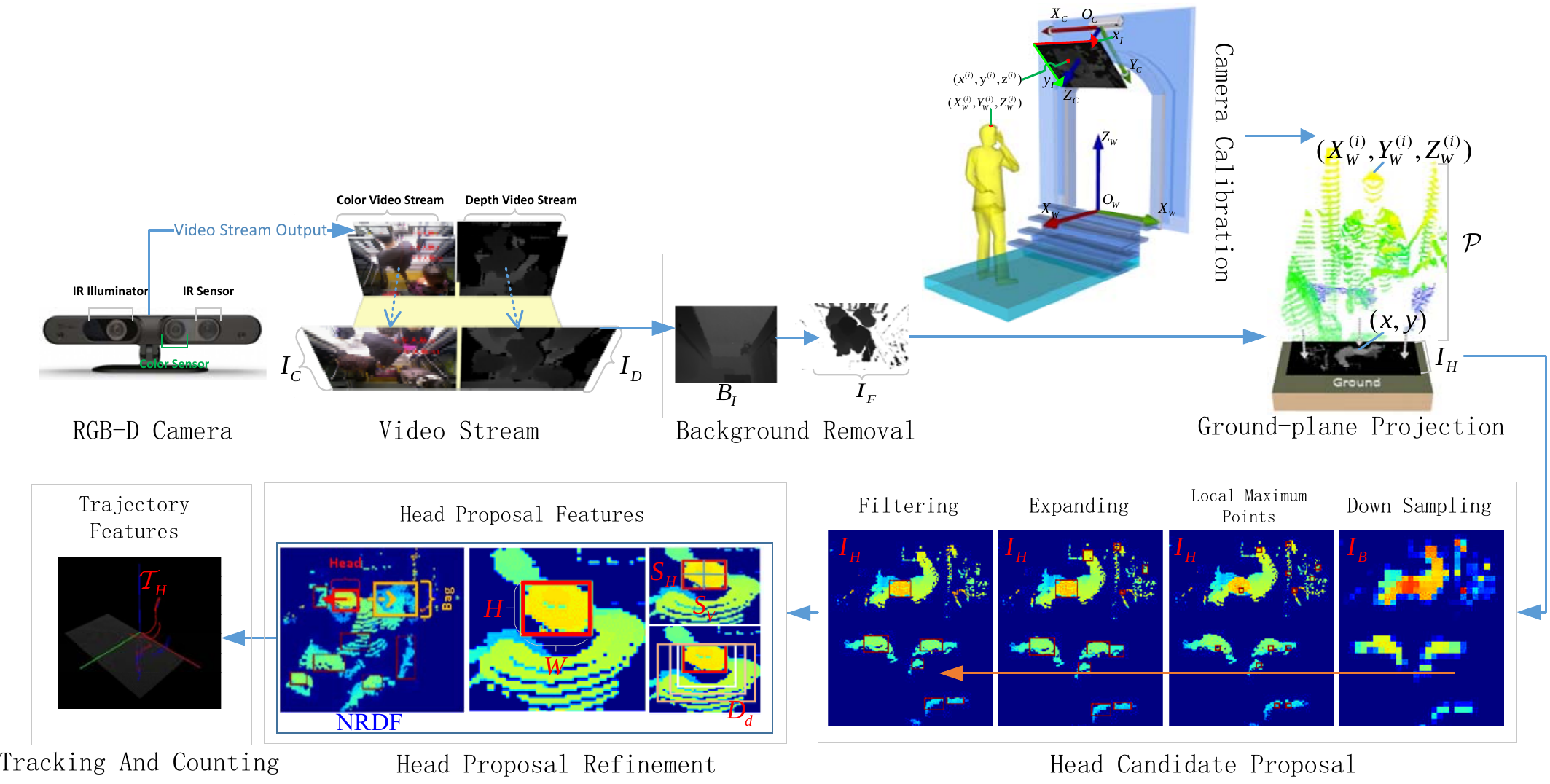}
\caption{Schematics of the proposed approach: The RGB-D camera provides color and depth video streams. The proposed method uses a depth frame ($I_D$) to extract the foreground for which 3D-point cloud is computed. The point-cloud is orthogonally projected onto the ground plane to generate a height image. Multiple potential head locations in the height image are computed by analyzing object features. These potential head proposals are refined and tracked continuously to count people entering or exiting a bus door.}
\label{fig:pipeline}
\vspace{-3mm}
\end{figure*}

Figure \ref{fig:pipeline} illustrates the pipeline of the  proposed method. Our approach assumes the setting where the camera is mounted on top of the area to be monitored (see `Camera Calibration' in the figure). This is the most common setting in scenarios like bus doors, corridors, entrances/exits to market places etc. After retrieving the depth video stream from the input, we  subtract the scene background using a proposed procedure for the real-time performance of our approach. A 3D-point cloud of the foreground is computed from the depth information and then re-projected orthogonally onto the ground plane for effective segmentation. For each video frame, we analyze its projected height image for the presence of potential human heads while employing a 3D-human model to refine those proposals.
The refined proposals are tracked to compute the head trajectories that are further refined and continuously monitored in our approach to count people entering or leaving the buses. To achieve our objective, we also introduce two discriminative feature vectors for head detection in height images and trajectory tracking in frame sequences. Our approach is evaluated using PCDS, and achieves up to 92\% accuracy for the real-world bus videos while enabling processing at 45 fps on a relatively less powerful 1.7GHz Intel processor with 2GB RAM. These results are significant since people counting is a challenging problem and our method can achieve real-time performance in practical conditions with minimal computational resources. 


This paper is organized as follows. The related work is reviewed in Section \ref{sec:RW} followed by Section \ref{sec:D} which describe the published dataset named PCDS. In Section \ref{sec:PA}, we introduce the proposed approach for people counting in cluttered environment. The experiments and results are provided in Section \ref{sec:E}. In Section \ref{sec:C}, we draw the conclusion.

\section{Related Work} 
\label{sec:RW}
The problem of people counting is often seen from two different perspectives: (a) Region of Interest (ROI) counting \cite{Chen2014, Chan2012a, Zhang2015, Idrees2013}, and (b) Line of Interest (LOI) counting \cite{Cong2009}. The former deals with counting  people in specific regions (e.g. in playgrounds), whereas the latter aims at counting the number of people `passing through' a certain region (e.g. through doorways). This work deals with the LOI counting. 
Many methods for LOI counting have been proposed, which can be divided into three major categories: 1) regression-based methods, 2) clustering-based methods, and 3) detection-based methods. 
Below, we review literature under these categories with emphasis on detection-based methods because of their close relevance to the proposed approach.

\vspace{-2mm}
\subsection{Regression-based Methods}
The main objective of the regression-based methods is to learn a regression function as the representation of changes in a scene which indicates passing of a pedestrian.
Under the paradigm of regression-based approach, Barandiaran et al.~\cite{Barandiaran2008a} used a single RGB camera to count people by the state change of virtual counting lines.
Pizzo et al.~\cite{DelPizzo2016a} proposed a method which divides the detection region into stripes and counts people by monitoring the change of state for these stripes without people head detection and object tracking steps.
Fradi et al.~\cite{Fradi2012} used Gaussian Mixture Model (GMM) to extract the foreground and used Gaussian Process regression to learn the correspondence between frame-wise features and the number of persons.
Benabbas et al.~\cite{Benabbas2010a} proposed a method which accumulates image slices and estimates the optical flow. They applied  a linear regression model to blob features which are  extracted by an on-line blob detector to get the position, velocity, and orientation of the  pedestrian.
Cong et al.~\cite{Cong2009a} estimated the number of pedestrians passing through a line by quadratic regression with the number of weighted pixels and edges which are extracted from the flow velocity field.
Whereas useful, one common drawback of the above methods is that they place hard restrictions on camera installation angles and the scene itself, which compromises their practical value. 

\subsection{Clustering-based Methods}
Clustering based methods simultaneously track multiple features of objects e.g. key points or  people component, and subsequently count people by clustering feature trajectories. For instance, Antonini et al.~\cite{Antonini2006} clustered  trajectories of visual features and then used the number of clusters for counting people. Topkaya et al. \cite{Topkaya2014} used features based on spatial, color and temporal information and clustered the detected feature trajectories by Dirichlet Process Mixture Models (DPMMs)~\cite{Neal2000}. They used Gibbs sampling to estimate an arbitrary number of people or groups in their approach. Brostow et al.~\cite{Brostow2006a} proposed a method that first tracks   simple image features and then probabilistically groups them into clusters based on space-time proximity and trajectory coherence through the image space.
Rabaud et al. \cite{Rabaud2006} used KLT tracker~\cite{Lucas1981} to track feature points, and segmented the set of trajectories by a learned object descriptor.

\subsection{Detection-based Methods}
The approaches that fall under this category share a common sequential processing pipeline which goes as follows. First, foreground is extracted from the video stream, then the objects of interest are detected and tracked. The tracked trajectories are subsequently classified to count the objects of interest. The detection-based methods  can be further divided into three different groups based on the underlying data modalities, namely 1) RGB video methods, 2) Depth video methods, and 3)~Hybrid methods. We also include an additional  category in our review that includes approaches employing the fast emerging deep learning framework. 

\subsubsection{RGB video methods}
Using RGB videos is more popular in people counting literature because of easy availability of color video cameras. Zeng et al.~\cite{Zeng2010} detected head-shoulder patterns in RGB videos by combining  multilevel HOG features~\cite{Dalal2005a} with multilevel LBP features~\cite{Ojala2002}. They used PCA~\cite{Wold1987} to reduce the dimensionality of the multilevel HOG-LBP feature set, and finally tracked the head-shoulder patterns to count people.
Antic et al.~\cite{Antic2009} proposed a people segmentation, tracking, and counting system by using an overhead mounted camera. 
Garcia et al. \cite{Garcia2013} also developed an RGB system for counting people in supervised areas. Their method is based on finding heads of people by a circular pattern detector and tracking them using Kalman filter~\cite{Julier1997}. Their approach also performs the final counting using the tracked trajectories.
Chen et al.~\cite{Chen2012a} used a vertical RGB video-camera to count a crowd of moving people by segmenting the crowd based on the frame difference method~\cite{Chaohui2007}. Their approach extracts features to describe the individual patterns, and tracks the individuals for counting. 
Kurilkin et al. \cite{Kurilkin2016} compared different people detectors in their study. 

The  methods described above are likely to suffer from critical failures when the scenes become complicated due to shadows, light changes, compound objects, occlusion, and the presence of significant background texture. To alleviate these problems, researchers exploit stereo cameras which can provide the third dimension information. For instance, Terada et al.~\cite{Terada1999} proposed one of the first approaches for stereo camera based people counting in  RGB video regime. 
They detected people using max points, tracked them with template matching and  finally used the two measurements from the stereo vision for counting.
In a related approach, Kristoffersen et al. \cite{Kristoffersen2016a} used two thermal cameras to reconstruct 3D points and proposed an algorithm for pedestrian counting based on clustering and tracking of the 3D point clouds. However, in their approach, the cost of depth computation remains high, which makes it difficult to use the approach  in real-time with low computational power devices. 

\subsubsection{Depth video methods}
With the popularity of RGB-D cameras; such as Kinect V1/V2~\cite{Zhang2012b} and Prime-Sense~\cite{Freedman2010}, depth videos are also becoming popular in people counting applications. 
Zhang et al.~\cite{Zhang2012a} proposed to use a so-called `water filling method' to detect people and counting them by the virtual line in a depth image. 
Barandiaran et al.~\cite{Barandiaran2008a}, and later Pizzo et al.~\cite{DelPizzo2016a, Pizzo2015} proposed approaches that are based on detection without tracking. 
These approaches detect changes in scene states across a virtual line, where the scene is divided by multiple stripes. The state of the scene changes when people pass by, thereby enabling people counting. 
Rauter et al.~\cite{Rauter2013} introduced the Simplified Local Ternary Patterns (SLTP) that are used to describe a human head. They trained an SVM  using SLTP and tracked human heads with the nearest neighbor association methods.  
Vera et al.~\cite{Vera2016a} proposed a network of cameras to count people. They devised a head detection method based on morphology geodesic reconstruction \cite{najman1996geodesic} and performed tracking using the Hungarian algorithm~\cite{kuhn1955hungarian}. 
Their approach combines tracks generated by multiple cameras and the final count is based on the length of the combined track. 
Li et al.~\cite{Li2017} proposed an embedded framework for real-time top-view people counting. They used the Kinect camera and the Jetson TK1 board~\cite{Ukidave2015} to  detect human heads using the  water filling technique~\cite{Zhang2012a}. Their approach also uses the  nearest neighbor association method for tracking. 
Enrico Bondi et al.~\cite{Bondi2014} introduced a framework for real-time people counting which follows the sequence of background removal, head detection and tracking the projected heads. Whereas promising, their framework drastically performs in complicated scenarios where other head-like objects also appear in the scenes.

\subsubsection{Hybrid methods}
\label{sec:Hybrid}
Combining the advantages of RGB and depth data streams are well documented in related problems, e.g. action recognition~\cite{Rabaud2006}.
Therefore, few methods have also used the hybrid approach in people counting.
For instance, Gao et al. ~\cite{Gao2016} detected head candidates in depth videos by water filling method and refined these candidates by training an SVM classifier using  HOG features of the frame in RGB videos.
Their approach eventually generates a set of trajectories by the nearest distance between the head candidates and the previous tracks. 
Liu et al.~\cite{Liu2013}  also used RGB-D camera for detecting people. Their approach projects people into a virtual plane and trains an SVM classifier using   features that are used for detecting the upper body of humans. 
Zhang et al.~\cite{Zhang2016} proposed head detection by blob detection in depth frames and projected the blobs into the 3D space.
Their approach filters the candidate blobs for heads by physical constraints and employs the histogram of multi-order depth gradient (HMDG) features and joint histogram of color and height (JHCH) features to train an SVM classifier.
The trained SVM is used to classify the candidate blobs as heads. However, their approach remains sensitive to occlusion.


\subsubsection{Deep Learning-based Methods}
Recently, Deep Learning~\cite{jia2014caffe,LeCun2015,Abadi2016} has demonstrated great success in object detection  and classification tasks~\cite{Cao2017}, which has also motivated researchers to employ it for the problem of people counting. For instance, Liu et al.~\cite{Liu2017} proposed a people counting system based on  Convolutional Neural Network (CNN)~\cite{Ciregan2012a} and  Spatio-Temporal Context (STC)~model~\cite{zhang2014fast}.  The CNN model is used to detect  people whereas the STC model is used to track  heads of moving people. Similarly, Wei et al.~\cite{Wei2017} proposed a framework based on supervised learning. They extracted spatio-temporal multi-features by joining super-pixel based multi-appearance features and multi-motion features, and then fused the multi-features with the features extracted from the VGG-16 model~\cite{Simonyan2014}. 

\begin{figure*}[tb]
\centering
\includegraphics[width=5in]{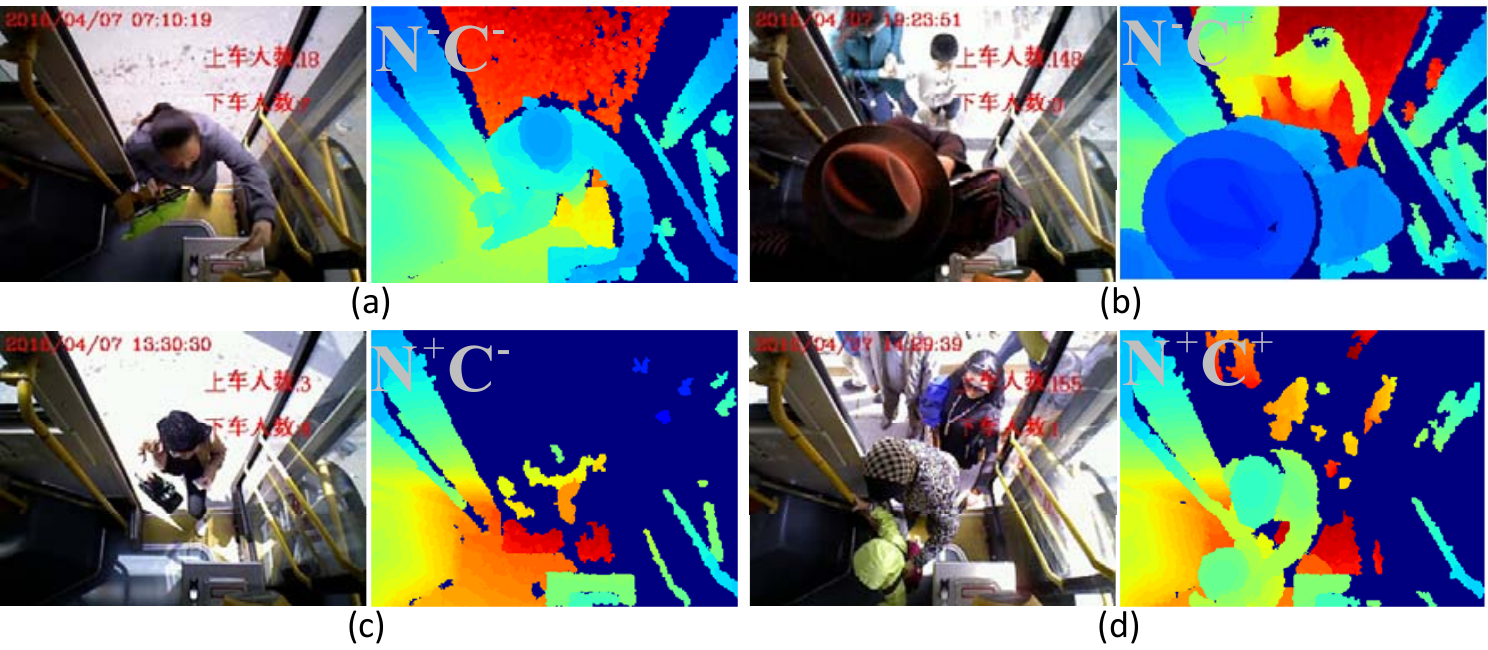}
\caption{Representative RGB and  depth images of different types of scenarios in PCDS: (a) Normal one-person entry. (b) A person wearing a large hat and the other person holding a child's hand. (c) Noisy sensor data. (d) Multiple people queuing with partial occlusion. }
\label{fig:datasetTypes}
\end{figure*}

\section{People Counting Dataset (PCDS)} \label{sec:D}
In this Section, we present the \textbf{P}eople \textbf{C}ounting \textbf{D}ata\textbf{S}et ($\rm{PCDS}$) introduced for the problem of people  counting in real-world conditions. 
The dataset is publicly available for download at~\url{https://github.com/shijieS/people-counting-dataset.git}. 

\subsection{Settings and Data Taxonomy}
\label{sec:PCDS_Tax}
The data consists of videos of bus-door scenes recorded using  Kinect V1 camera~\cite{Zhang2012b}. The camera is mounted on the ceiling of (front/back) doors of different buses, and captures people entering or exiting through the doors. Figure \ref{fig:datasetTypes} shows four representative scenes from the dataset. Due to the real-world scenarios, complexity of the data is  apparent from the figure. 
In comparison to the existing related datasets~\cite{DelPizzo2016a, Zhang2012a}, videos in PCDS are recorded by the camera with a pitch angle that is not necessarily vertical to the ground plane.

\begin{figure}[t]
\centering
\includegraphics[width=3in]{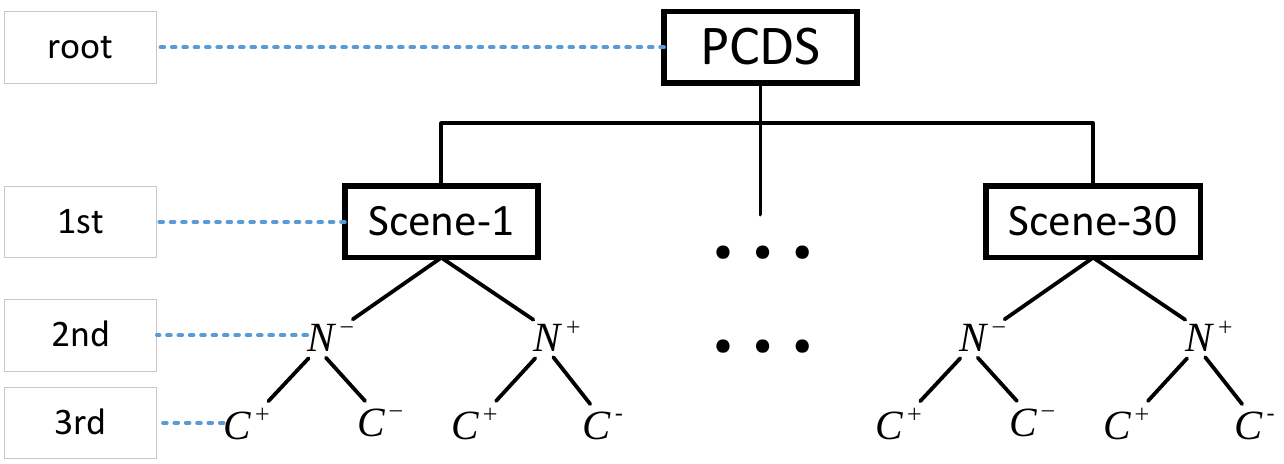}
\caption{Structure of the \textbf{P}eople \textbf{C}ounting \textbf{D}ata\textbf{S}et ($\rm{PCDS}$): The root directory contains 30 scenes, with two subdirectories each. Each subdirectory either contains noisy scenes (denoted by $N^+$) or clean/normal scenes (denoted by $N^-$). The 3rd-level of directories contains crowded (i.e. $C^+$) or un-crowded (i.e. $C^-$) scenes.}
\label{fig:videoClasses}
\end{figure}

We divide the videos in the dataset based on the bus route numbers. The dataset is recorded for three different bus routes, namely No.~25, No.~301 and No.~106 in the cities of Xi'An, XiNing  and YinChuan,  respectively in China. The data samples cover all the bus stops in the complete circuit route of the buses. For No.~25, the videos have been collected on 6 different days. For No.~301 and No.~106, the number of days are 5 and 4 respectively.
For each day, we collected data for the front door as well as the back door. Thus, in total, there are 30 different scenes in our dataset. 
We can further sub-categorize the videos of these scenes based on their noise level and crowd in the scene.
In the above mentioned URL for the dataset, we organize the dataset according to these measures. 
In Fig.~\ref{fig:videoClasses}, we provide the folder structure of the proposed dataset. 
We denote the noisy/clean scenes with $N^+/N^-$, and crowded/un-crowded scenes by $C^+/C^-$. As an example, a noisy-crowded scene is denoted by $N^+C^+$ according to the adopted notation.

The rationale of dividing the dataset into noisy and clean videos is that Kinect V1 camera is sensitive to illumination conditions. For strong illumination, there is often noise in the videos, as can also be observed in Fig.~\ref{fig:datasetTypes}. The videos in our dataset are mainly recorded in either direct sunlight or diffused sunlight, resulting in a natural division of corresponding levels of noise.
Similarly, the division of videos according to congestion in the scenes is also natural.
During rush hours, multiple people are generally passing through the bus doors. On the other hand, sequential entry with clear separation between people is observed during normal conditions.

In light of the division provided in Fig.~\ref{fig:videoClasses}, one can expect the following from the four possible sub-categories for  each scene in the proposed dataset:
\begin{itemize}
\item $N^+C^+$: Videos are captured in strong sunlight during rush hours, with multiple people attempting to enter/exit the bus at once.
\item $N^+C^-$: Videos recorded in sharp sunlight  during normal hours where people are entering/exiting bus doors in a more sequential manner.
\item $N^-C^+$: The recording is performed with mild sunlight but in crowded situations.
\item $N^-C^-$: The recording is done in mild/diffused sunlight with sequential entry/exit of people through the doors. 
\end{itemize}
Table.~\ref{tab:numberOfDataset} summarizes the number of people entering and exiting the bus doors for each sub-category. 

\begin{table}[t]
\centering
\caption{The number of people in $\rm{PCDS}$}
\label{tab:numberOfDataset}
\begin{tabular}{ccccc}
\hline
        &    $N^-C^-$    &    $N^-C^+$       & $N^+C^-$    &     $N^+C^+$     \\ \hline
Entering&    2,704        &    5,427          & 616        &     937          \\
Exiting    &     2,760        &    6,647        & 668        &   1,149        \\ \hline
Total    &    5,464        &    12,074        & 1,284        &   2,086        \\
\hline
\end{tabular}
\end{table}

\subsection{Video Information}
In Table~\ref{tab:basicVideoInformation}, we summarize the basic attributes of the videos in our dataset. We note that, these video attributes along the camera parameter details are also provided in each folder of the dataset. Moreover, the total number of people passing through the doors is also provided as the ground truth.  
We also provide RGB videos along the depth videos that can be used for verification purposes. However, we emphasize that the depth modality is more useful for the problem of people counting in the real-world conditions because of its robustness to e.g, illumination conditions and shadows. 

\begin{table}[t]
  \centering
  \caption{Video attributes}
    \begin{tabular}{ccccc}\hline
           Type       & fps   & Resolution         & Channels    & Count  \\\hline
    RGB video & 25    & $320 \times 240$     & 3         & 4,689    \\
    Depth video & 25    & $320 \times 240$     & 1         & 4,689  \\\hline
    \end{tabular}%
  \label{tab:basicVideoInformation}%
\end{table}%

\section{Proposed Approach} \label{sec:PA}
\label{sec:Prop}
The schematics of the proposed approach for people counting is illustrated in Fig.~\ref{fig:pipeline}. Our method performs counting by analyzing the depth video frames retrieved from the RGB-D camera. The major steps involved in our approach are; {1)~removing the scene background, 2) re-projecting point cloud onto the ground plane, 3) generating candidate head proposals in the projected images, 4) refining those proposals, and 5)~tracking the trajectories of human heads for counting.} We provide details of each of these steps below. 





\subsection{Background Removal}
\label{sec:BR}

There are multiple techniques for background subtraction from RGB videos~\cite{Zeng2010}, \cite{Kaewtrakulpong2001},  \cite{Zivkovic2004}. 
However, depth videos are inherently different from RGB videos and such methods are not readily applicable to them.
Few methods for background removal from depth videos also exist~\cite{Del-Blanco2014,Chacon-Murguia2016}.
However, those techniques are generally computationally expensive, which makes them unsuitable for our real-time application. Moreover, such  methods were also found to be unsuitable for handling the noise in PCDS resulting from the real-world conditions. 
Therefore, we develop our own method for efficient background subtraction  from depth videos for people counting scenarios, such that the results  also  remain robust to noise in the real-world data.




In our settings, a depth frame $I_D\in \mathbb{R}^{H \times W}$ is a matrix, with its each element representing the distance of a point in the real-world to the camera sensor. 
For a camera mounted on top of the area to be monitored (as in PCDS), one can expect that the farthest points in the scene would generally belong to the background.
Based on this simple intuition, we develop a `farthest background model' $B_I\in \mathbb{R}^{H \times W}$ of dynamic scenes that enables automatic estimation of the background on-the-fly. A major advantage of such an approach is that it can be readily used for any scene without the need of calibration for the background.

We compute $B_I$ as a map of the largest distances appearing in the sequences of depth frames, while accounting for the possible noise accumulation. 
To ensure that effective $B_I$ is available for each video frame, we take the help of two intermediate models $B_c$ and $B_{2c}$, where `$c$' stands for cache. We initialize $B_I$ and $B_c$ with $I_D$ at the start of the video stream ($B_{2c}$ is initialized later, see below). 
For an input frame sequence, we update $B_c$ at every frame as follows:
\begin{equation}
\label{eq:backgroundUpdateCache}
B_c^{t} = \mathop{max}\{I_D^{t}, B_{c}^{t-1}\},
\end{equation}
where the superscript `$t$' denotes the current frame and  ${t-1}$ indicates the previous frame. The $max\{.\}$ operation is performed element-wise. After every $n_c$ frames, we update $B_I$ by assigning it  the values of $B_c$. 

It is easy to see that under the above mentioned strategy, any large distance values in $B_c$ resulting from noise at any stage can eventually  get stored in $B_I$. To cater for this problem, we separately initialize $B_{2c}$ with $I_D$ just after $B_I$ is updated (i.e. after $n_c$ frames), and keep updating it with every frame as follows:
\begin{equation}
\label{eq:backgroundUpdate2Cache}
B_{2c}^{t} = \mathop{max}\{I_D^{t}, B_{2c}^{t-1}\}.
\end{equation}
We update $B_c$ as well as $B_I$ with $B_{2c}$ after each $n_{2c}$,  whereas we impose that  $n_{2c} - n_c \neq 0$ to ensure that the update of $B_I$ under $B_c$ and $B_{2c}$ is asynchronous. This strategy entails that a maximum value once entered in $B_I$ as a result of noise can be replaced by the correct smaller value in the later frames. For computational purpose, we also constrain $n_{2c} > 2n_c$. As a result of the asynchronous updates with intermediate models, effective $B_I$ remains available for each frame. We use this farthest background model to extract the foreground $I_F$ at each frame as follows:   
\begin{equation}
\label{eq:foregroundImage}
I_F^{t}(u,v) = \left\lbrace
\begin{array}{ll}
0 ,~~~~~~~~~~~|B_I^{t}(u,v) - I_D^{t}(u,v)| < \delta_{dis}\\
I_D^{t}(u,v),~~~\text{otherwise}
\end{array}\right.
\end{equation}
where $B_I^{t}(u,v)$ is the pixel value at $(u,v)$ position of $B_I^{t}$, $I_D^{t}(u,v)$ is the pixel value at $(u,v)$ position of $I_D^{t}$, and $\delta_{dis}$ denotes the threshold parameter for our approach.

\newcommand{\pw}[2]{#1_W^{(#2)}}
\newcommand{\pc}[2]{#1_C^{(#2)}}
\subsection{Reprojection}
\label{sec:reprojection}
\begin{figure}[t]
\centering
\includegraphics[height=2.0in]{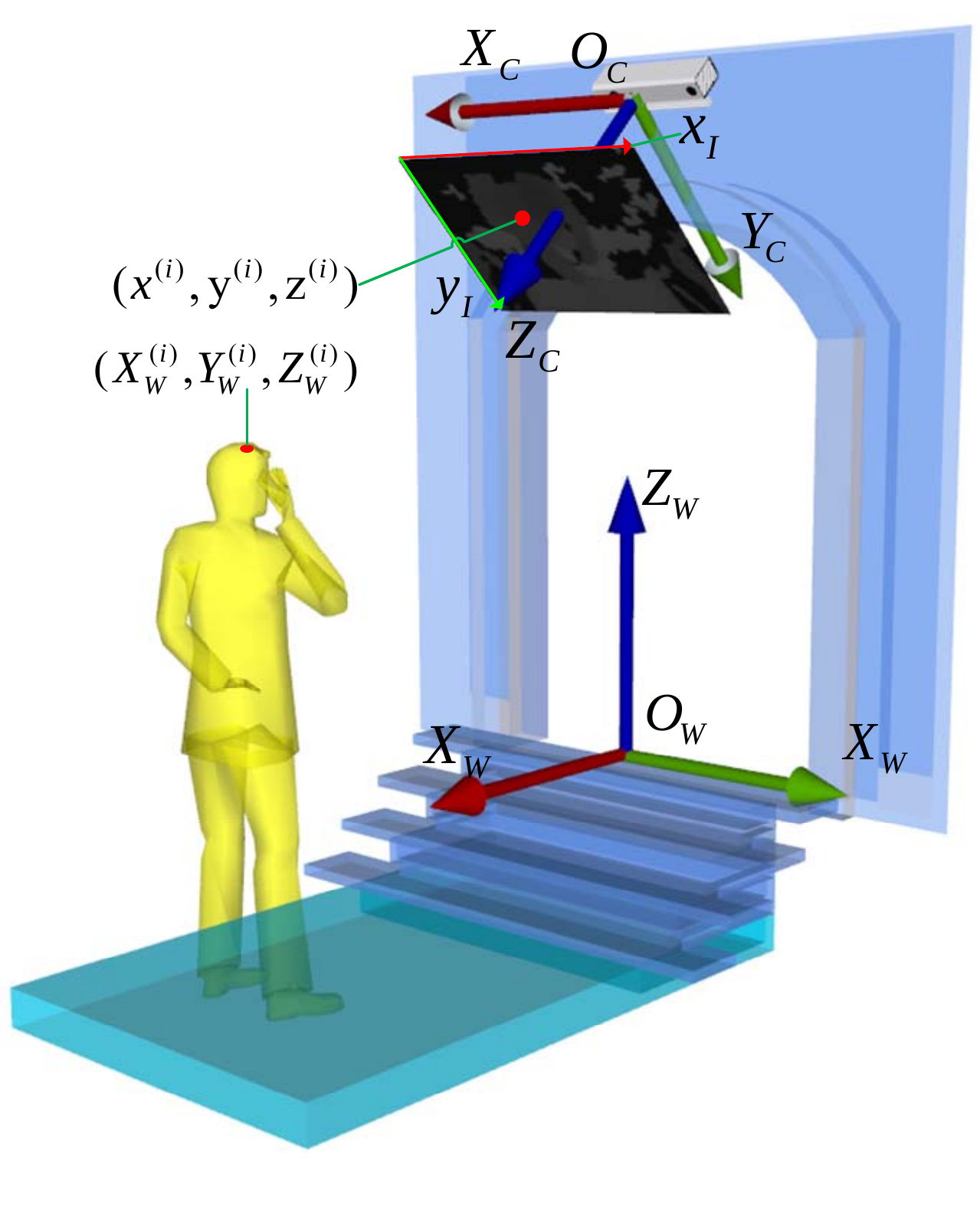}
\caption{Camera mounted position and camera calibration: The RGB-D camera is mounted on top of the scene and has a pitch angle. The camera coordinate frame is shown by $(X_C, Y_C, Z_C, O_C)$, the world frame is given by $(X_W, Y_W, Z_W, O_W)$, and $(x_I, y_I)$ represent image coordinates. $(x^{(i)}, y^{(i)}, z^{(i)})$ denotes the i-th point in image coordinates where $z^{(i)}$ is the pixel value. $(X_W^{(i)}, Y_W^{(i)}, Z_W^{(i)})$ is the i-th point in the world coordinates which corresponds to $(x^{(i)}, y^{(i)}, z^{(i)})$. }
\label{fig:cameraCalibration}
\end{figure}

Generally, cameras used for counting people are installed with non-zero pitch angle, e.g. see Fig.~\ref{fig:cameraCalibration}. The camera perspective often causes occlusion and overlap in the depth maps of people, which adds to the complexity of counting problem.
The objective of ``reprojection'' stage is to remove the perspective distortions so that individuals become well separated in the reprojected depth frames. 
For that purpose, we first construct a 3D point cloud from a depth frame of the camera and then re-project it normally onto the ground plane to obtain a normalized depth image. We present details of the reprojection procedure below.


First, we convert the foreground image $I_F$ into 3D points in the camera coordinates. 
For every pixel in $I_F$, we recover its 3D point as follows: 
\begin{equation}
\label{eq:translateCameraCoordinate}
\left\{
\begin{matrix}
X_C =& \frac{u - c_x}{f_x}\cdot I_F(u,v) \\
Y_C =& \frac{v - c_y}{f_y}\cdot I_F(u,v) \\
Z_C =& I_F(u,v),
\end{matrix}
\right.
\end{equation}
where $(f_x, f_y)$ denote the camera focal length, $(c_x, c_y)$ is the camera principal point, $I_F(u,v)$ is the pixel value at the position $(u,v)$ in $I_F$, and $X_C, Y_C, Z_C$ are the recovered 3D point coordinates. 


For projecting points onto the ground, we must first convert 3D points in the camera coordinates to the world coordinates.
Let us denote the world coordinate frame by $\{X_W, Y_W, Z_W, O_W\}$. We fix this frame directly below the camera coordinate reference frame, as shown in Fig.~\ref{fig:cameraCalibration}. 
To perform the transformation between the coordinate frames, we  compute the homogeneous transformation matrix (${\bf T} \in \mathbb{R}^{4\times4}$) based on the extrinsic parameters of the camera. To that end, we first identify $N$ points in a depth frame acquired by the camera and physically measure the corresponding points in the world coordinates. The  following optimization problem is then solved using the least squares approach~\cite{Bjorck1996}:
\begin{align}
<{\bf T}> = \min_{{\bf T}} ||{\bf P}_W - {\bf T}~~{\bf P}_C||_F^2,
\end{align}
where ${\bf P}_W \in \mathbb R^{4\times N}$ contains $N$ points arranged  as its columns in the world coordinates, and ${\bf P}_C \in \mathbb R^{4\times N}$ contains the corresponding points in the camera coordinates. The last row of these matrices consist of 1s.  For a unique solution, we constrain $N > 4$ in our measurements.

Note that, estimation of ${\bf T}$ is an off-line process in our approach and it is  performed only once for calibration.
Using the matrix ${\bf T}$ we eventually transform all points in $I_F$ to a 3D point cloud in the word coordinates. We then project this point cloud normally onto the ground plane. Intuitively, multiple points in the 3D point cloud can be mapped to the same point on 2D ground plane. 
In our approach, we only store the 2D mappings of the highest points in the 3D point cloud. 
Concretely, for the points $(\pw{X}{i}, \pw{Y}{i}, \pw{Z}{i}), \forall i$ in the 3D point cloud, we compute a 2D ground plane projection $I_H{(x,y)}$  as follows:
\begin{equation}
\label{eq:projectPointCloud}
\left\{
\begin{matrix}
\mathcal{Z}^{(x,y)} =& \{\pw{Z}{i}|\pw{X}{i}=x \wedge \pw{Y}{i}=y, \forall i\} \\
I_H{(x,y)} =& \max(\mathcal{Z}^{(x,y)})
\end{matrix}\right.
\end{equation}
where $(x,y)$ indexes points in the 2D plane. Henceforth, we refer to $I_H$ as the ``height image'' because each point/pixel in this image represents the highest point in the corresponding 3D point cloud. 

\begin{figure}[t]
\centering
\includegraphics[height=2.5in]{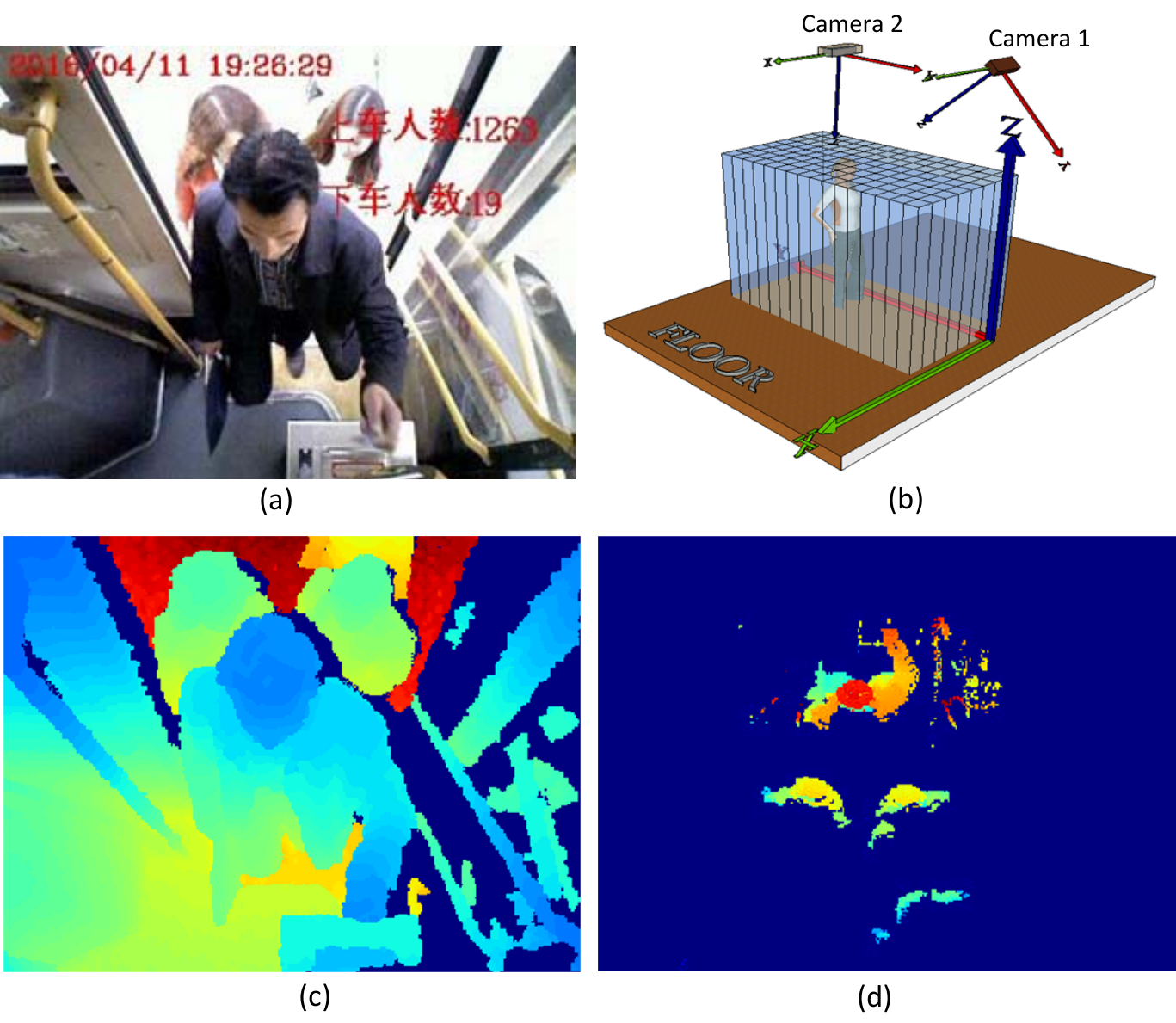}
\caption{Effects of reprojection: a) Color frame. b) Illustration of change in camera perspective from Camera 1 to Camera 2 by reprojection. c) Depth frame obtained by camera. d) Depth image constructed by reprojection.}
\label{fig:projection}
\end{figure}

The effects of reprojection can be understood as acquiring the depth image by `Camera 2' instead of `Camera 1' in Fig.~\ref{fig:projection}b.
Looking at the scene exactly from the top, Camera 2 is able to separate the individuals in the scene very well. This is clear from  Fig.~\ref{fig:projection}d when we compare it with the camera captured depth frame (with background) in Fig.~\ref{fig:projection}c.
The figure clearly demonstrates the benefits of the reprojection step in the proposed approach. 

\subsection{Candidate Head Proposals} \label{sec:locationgHead}
Although the reprojected height images separate individuals well, the loss of information due to occlusions in the original depth frames can not be recovered from these images.
Due to  their height, human heads suffer the least from the occlusions caused by the camera perspective.
Therefore, instead of tracking the complete human body to count people, we focus on reliable localization of human heads in the height images and eventually use head trajectories for people counting. 


\begin{figure}[t]
\centering
\includegraphics[height=2.0in]{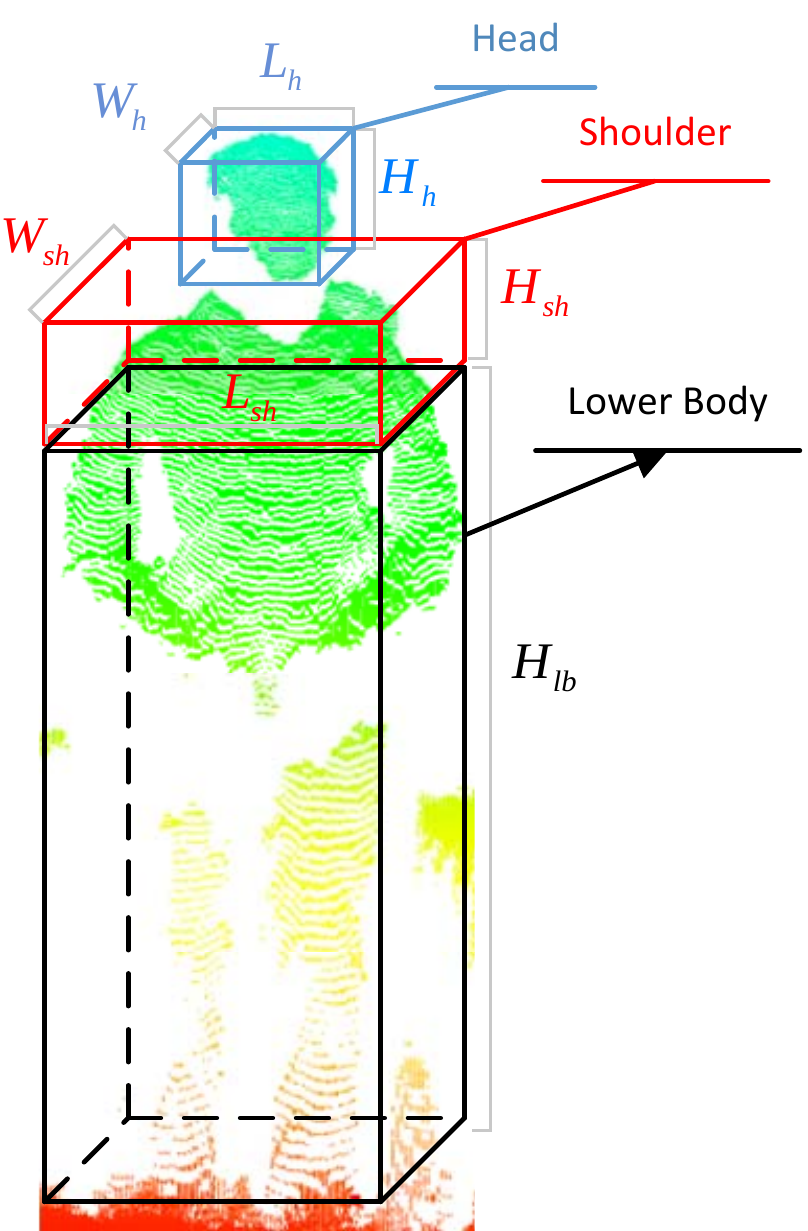}
\caption{Division of human body used: A body is divided into `Head', `Shoulders' and `Lower body' parts and a cuboid is placed on each part. Practical thresholds are placed on the dimensions of each cuboid to localize the corresponding part in the height image.}
\label{fig:3PartOfPeople}
\end{figure}

To locate human heads in the reprojected images, we exploit our prior knowledge about a human body.
We make use of a 3D human model, shown in Fig.~\ref{fig:3PartOfPeople}, by dividing the body into three major parts, namely; \emph{Head}, \emph{Shoulders}, and \emph{Lower body}.
We place a cuboid on each of these parts that represents the volume where the body part is most likely to be located in the 3D space. 
For instance, we expect the \emph{head} to be located in the blue cuboid of dimensions $W_h, H_h, L_h$ in the figure. The exact dimensions of the cuboid  would vary from person to person.
Therefore, we empirically place minimum and maximum thresholds on these dimensions in our approach.
The used thresholds are summarized in Table~\ref{tab:parameterOfHumanModel}.
Our intuition is that we can locate the corresponding body parts of individuals in the height image using the human model. Therefore, the thresholds in the table cover  reasonably large ranges to account for the variability in human sizes in height images.

\begin{table}[t]
  \centering
  \caption{Thresholds of each part of human model.}
    \begin{tabular}{cccccccc}
    	  \hline
          & & \multicolumn{2}{c}{\textbf{Head}} & \multicolumn{2}{c}{\textbf{Shoulder}} & \multicolumn{2}{c}{\textbf{Lower body}} \\ 
          & & min   & max   & min   & max   & min   & max \\ \hline
    $L$   & pixels & 10    & 25    & 25    & 60    & / 	  &   / \\
    $W$   & pixels & 10    & 25    & 10    & 20    & /	  &   / \\
    $H$   & cm & 15    & 30    & 10    & 30    & 60    & 170 \\ \hline
    \end{tabular}%
  \label{tab:parameterOfHumanModel}%
\end{table}%


With the help of underlying human model, we generate candidate proposals about the  human heads possibly present in the height image by sequentially performing the following steps. 1) Down sampling the height image, 2) computing the local maxima in the down sampled image, 3) expanding the local maximum points, and 4) filtering the expanded areas. Below, we describe each of these steps in detail.


\subsubsection{Down sampling}
\begin{figure}[t]
\centering
\includegraphics[width=3in]{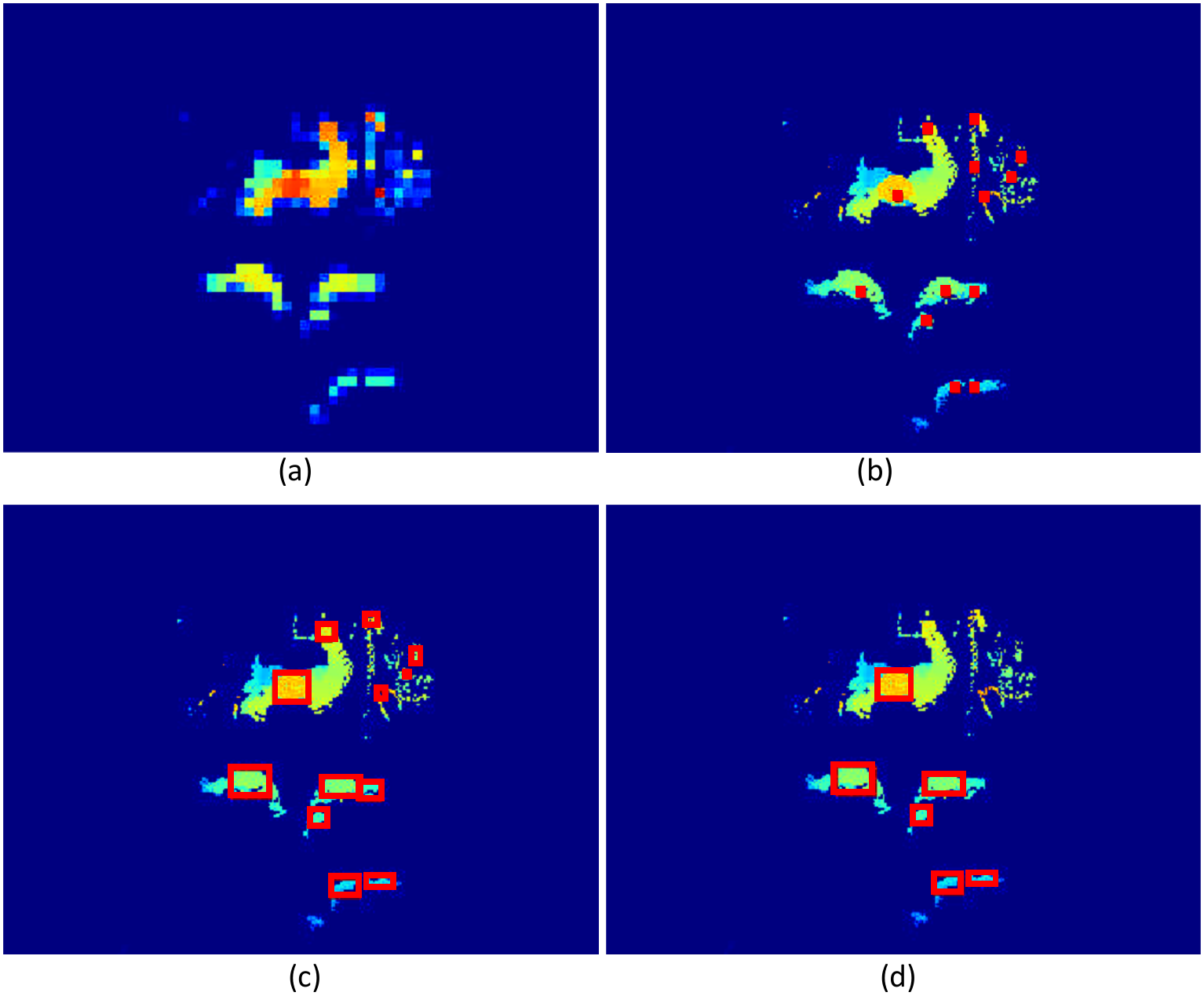}
\caption{Intermediate results for generating head proposals: a) Down sampled height image. b) Local maximum point in original height image. c) Expanding local maximum points. d) Filtering the local maximum areas.}
\label{fig:resultLocatingHead}
\end{figure}

Recall that our aim is to develop a real-time method that can deal with the real-world noise.
To reduce computations and mitigate the adverse effects of noise in this step, we first down sample the height image $I_H$ by averaging its $w_b \times w_b$ dimensional disjoint patches. As a result, we get an image $I_B$ with its pixel at $(x, y)$ location computed as follows: 
\begin{equation}
\label{eq:downSampling}
I_B{(x,y)} = \frac{\mathop{\sum}\limits_{u=w_bx}^{w_bx+w_b}\mathop{\sum}\limits_{v=w_by}^{w_by+w_b}I_H(u,v)}{w_b^2},
\end{equation}
where, $(u, v)$ denotes a pixel location in $I_H$. For the height image shown in Fig.~\ref{fig:projection}d, the resulting down sampled image $I_B$ is illustrated in Fig.~\ref{fig:resultLocatingHead}a.

\subsubsection{Local Maximum Point Computation}
Intuitively, the pixels corresponding to human heads are more likely to have the largest values in  height images. This property is also well preserved in the down sampled image $I_B$, as can be seen in Fig.~\ref{fig:resultLocatingHead}a.
Thus, to locate the areas that can potentially belong to human heads in $I_B$, we adopt a simple strategy of identifying a set $\mathcal{C}_b$ of the pixels in $I_B$ that contain the maximum values in their 8-connected pixels. These pixels are then used to identify the local maximum points in the original height image. Note that, the $i^{\text{th}}$ element of $\mathcal{C}_b$, i.e. $\mathcal{C}_b^i$ is computed as the mean of a set of pixels in the height image. We represent the set of the desired maximum pixels in $I_H$ as $\mathcal C_H$, and compute the $j^{\text{th}}$ element of that set, i.e.~$\mathcal{C}_H^j$ as follows:
\begin{equation}
\label{eq:translate2HeightImage}
\mathcal{C}_H^j = \mathop{max} \{\text{pixels in}~I_H~\text{corresponding to}~\mathcal{C}_b^i \}.
\end{equation}
As a result of this operation, we are able to efficiently identify the local maximum points in our height image. Fig.~\ref{fig:resultLocatingHead}b illustrates the computed points from the corresponding down sampled image in  Fig.~\ref{fig:resultLocatingHead}a.


\subsubsection{Expanding Local Maximum Points}
A local maximum point in $I_H$ may or may not belong to a human head. Therefore, we must analyze the local vicinity of the maximum point and compare it with our human model to ascertain that the point is indeed located on a human head.
We adapt the seed fill method~\cite{torbert2016applied} to expand the local maximum points into rectangles such that the object bounded by each rectangle can be compared with the human model. The procedure for expanding the local maximum point is given as Algorithm~\ref{alg:expandLocalMaximumPoints}.


\begin{algorithm}[t]
 \caption{$Expanding Local Maximum Points$}
 \label{alg:expandLocalMaximumPoints}
 \begin{algorithmic}[1]
 \renewcommand{\algorithmicrequire}{\textbf{Input:}}
 \renewcommand{\algorithmicensure}{\textbf{Output:}}
 \REQUIRE~~\\
 $I_H$: height image. $\mathcal{C}_H$: local maximum point set.\\
 $\delta_h$: expanding threshold. $W_{max}$: maximum head width.\\
 $L_{max}$: maximum head length.
 \ENSURE~~\\
 $\sset{E}_H$: the set of expanded rectangles.
 \renewcommand{\algorithmicensure}{\textbf{Initialize:}}
  \ENSURE~~\\
 $H \gets rows(I_H); $ \% height of the image.
 \\ $W \gets columns(I_H);$  \% width of the image.
  \FOR{\textbf{each pixel} $(x_0, y_0) \in \mathcal{C}_H$}
  \STATE $l \gets W; r \gets 0; t \gets H; b \gets 0$;
  \STATE $\mathcal{C}_0 \gets \{(x_0, y_0)\}; \mathcal{C}_{s} \gets \mathcal{C}_0;$
  \FOR{\textbf{each pixel} $(u, v) \in \mathcal{C}_0$}
  \STATE $\mathcal{N}_P$ is 8-connected pixels of $(u, v)$ 
  \FOR{\textbf{each pixel} $(x, y) \in \mathcal{C}_0$}
  \IF {$(x,y) \notin \mathcal{C}_{s}$ \AND $|I_H(x_0,y_0) - I_H(x,y)| \leq \delta_h$\\
  \AND $r-l_t \leq W_{max} $ \AND $b-t_t \leq L_{max}$}
  \STATE $l_t\gets \mathop{min}(x,l); r_t\gets \mathop{max}(x,r);$
  \STATE $t_t\gets \mathop{min}(y,t); b_t\gets \mathop{max}(y,b);$
  \STATE $\mathcal{C}_{s} \gets \mathcal{C}_{s} \cup \{(x,y)\}$
  \STATE $\mathcal{C}_0 \gets \mathcal{C}_0 \cup \{(x,y)\}$
  \ENDIF
  \ENDFOR
  \STATE $\mathcal{C}_0 \gets \mathcal{C}_0-\{(u, v)\}$
  \ENDFOR
  \STATE $\sset{E}_H \gets \sset{E}_H \cup \{(l, r, t, b)\}$
  \ENDFOR
  \RETURN $\sset{E}_H$
 \end{algorithmic}
 \end{algorithm}

Along the height image $I_H$ and the set of local maximum points $\mathcal C_H$, the algorithm requires the maximum allowable height and width of a head (from Table~\ref{tab:parameterOfHumanModel}) as the input. It also uses an expanding threshold $\delta_h$ as an input parameter, that restricts the expanded rectangles to contain object pixels with similar values. The algorithm eventually results in a set $\mathcal E_H$ that contains the expanded rectangles as its elements.
The main iteration of Algorithm~\ref{alg:expandLocalMaximumPoints} runs over each element of $\mathcal C_H$, that are called \emph{seeds} in the context of seed fill method~\cite{torbert2016applied}. The first inner `for loop' (lines 4-15 in the algorithm) performs the actual expansion process. It first identifies the 8-connected neighborhood of a considered pixel (line 5) and then iterates over each of the neighboring pixels (line 6-13) to evaluate the condition given on line 7 of the algorithm. If the condition is satisfied, the sets $\mathcal C_s$ and $\mathcal C_0$ are updated which are subsequently used in the outer loop.
The algorithm gradually expands a seed in $\mathcal C_H$ to a rectangle that is upper-bounded by the maximum dimensions $W_{max}$ and $L_{max}$ while ensuring that the pixel values in the expanded rectangles remain close to the seed value so that the rectangle only bounds a single object.

\subsubsection{Filtering Local Maximum Areas}
Despite capitalizing on the physical attributes of human body parts, we can still expect that few rectangles in $\sset{E}_H$ may not actually belong to human heads (see e.g.~Fig.~\ref{fig:resultLocatingHead}c).
Therefore, we further filter the computed rectangles using the human model.
In the filtration process, we also consider incomplete human heads resulting from occlusions.
To filter, we discard all the rectangles in $\sset{E}_H$ that do not satisfy the following condition:
\begin{equation}
\label{eq:conditionOfHead}
\left\{\begin{matrix}
\frac{L_{min}}{2} \leq L \leq L_{max}\\ 
\frac{W_{min}}{2} \leq W \leq W_{max},
\end{matrix}\right.
\end{equation}
where, $L$ and $W$ represent the length and width of a rectangle and the subscripts $min, max$ denote the minimum and maximum lengths allowed in Table~\ref{tab:parameterOfHumanModel} for `Head'. Notice that we reduced the minimum allowed values in Eq.~(\ref{eq:conditionOfHead}) by half. This is done to account for occlusions that can often cause the size of a head in our height image $I_H$ to reduce significantly.

\begin{figure}[t]
\centering
\includegraphics[width=3.3in]{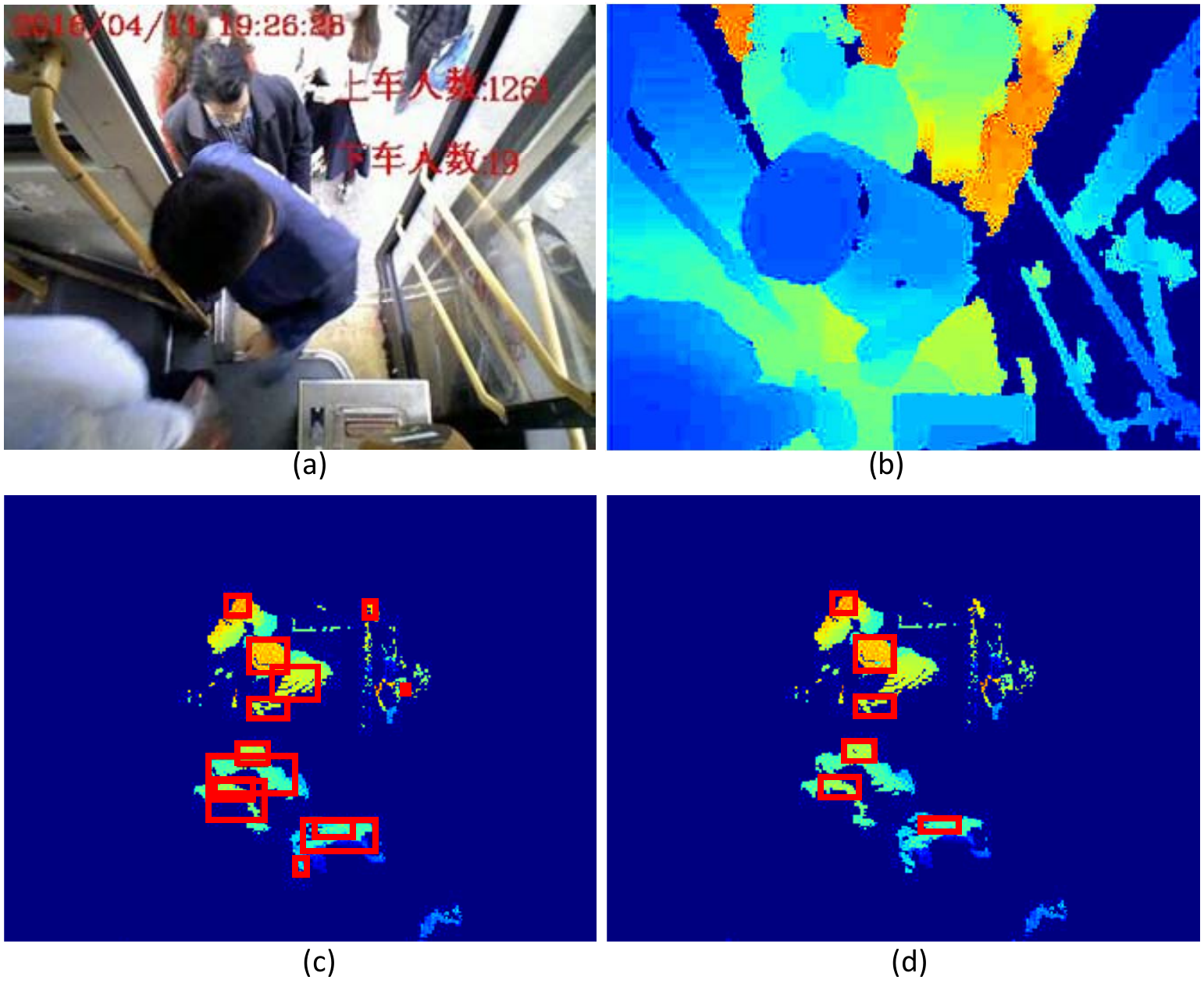}
\caption{Result of removing overlapped rectangles. a) color image, b) depth video frame, c) expanded local maximum areas, d) result of removing overlapped rectangles.}
\label{fig:problemOfFilterings}
\end{figure}

\subsection{Head Proposals Refinement}
\label{sec:HPR}
For the cameras installed on top of pathways, human heads in video frames rarely overlap in real scenes, as can be observed in Fig.~\ref{fig:problemOfFilterings}a-b. However, the set $\sset{E}_H$ may contain few overlapping rectangles (see Fig.~\ref{fig:problemOfFilterings}c), therefore we can further refine this set by discarding the overlapping rectangles.
To do that, we consider all the groups of overlapped rectangles in $\sset{E}_H$, and for each group, we store only the rectangle with the highest seed value and discard the remaining rectangles. 
We denote the refined set of rectangles by $\sset{F}_H$.
Fig.~\ref{fig:problemOfFilterings}d illustrates the result of this refinement.

The rectangles contained in the set $\sset{F}_H$ are highly likely to correspond to human heads in $I_H$, however it is still possible that some of those rectangles may actually belong to other objects in the scene.
Differentiating between a head and a non-head rectangle in $\sset{F}_H$ is a non-trivial task because occlusions and other factors, e.g.~presence of high round-shaped objects like bag-backs, can result in patterns in $I_H$ that are very similar to human heads.
We hypothesize that despite their close similarity with the human heads, the non-head objects can be automatically identified by analyzing their relevant features.
Hence, we design a compound discriminative feature that accounts for different relevant attributes of objects to classify them as `heads' and `non-heads'. We momentarily defer the discussion on the proposed feature to the text to follow. We extract the proposed features for the elements of $\sset{F}_H$ and train an SVM classifier over those features to further discard the rectangles that bound non-head objects. 

\begin{figure}[t]
\centering
\includegraphics[width=3.5in]{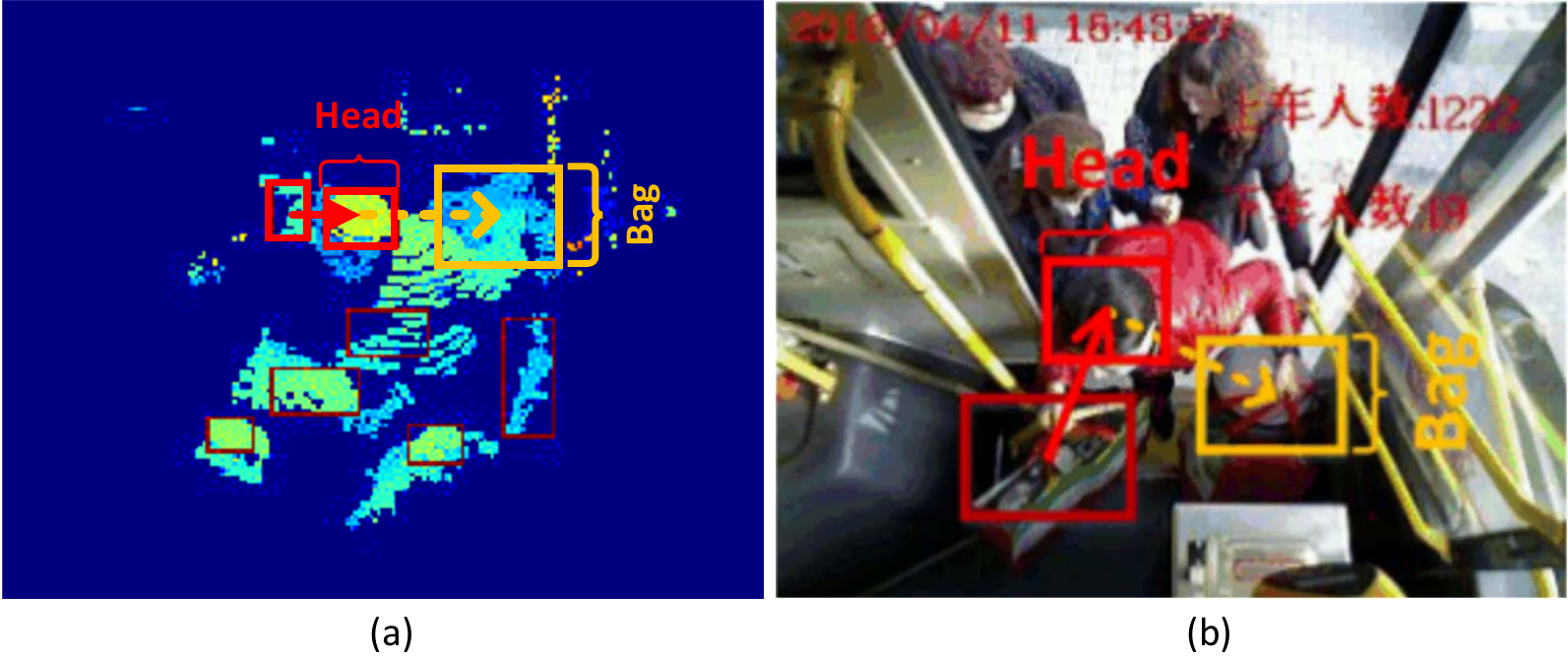}
\caption{Illustration of NRDF feature: a) The height image. b) The corresponding color video frame. There is a bag rectangle (yellow) in both images, whose nearest rectangle is the head (red). The yellow arrow is the NRDF feature of the bag rectangle and the red arrow is the NRDF feature of the head rectangle.}
\label{fig:nrcvFeatures}
\end{figure}

For the SVM training, we manually label each extracted feature for a rectangle as `head' or `non-head'. This off-line training is carried out only once in our approach on the training data. 
For the test frames, we similarly extract the features of head proposals and classify them as `heads' or `non-heads' using the trained SVM. The `non-heads' are discarded in further processing.
Our proposed compound feature vector is a concatenation of two major types of features that we call \emph{Basic Geometric Features} (BGF) and the \emph{Nearest Rectangle Difference Feature} (NRDF). The BGF itself is a combination of four different features explained below:

\begin{itemize}
\item \emph{Shape Feature} $(H_r,W_r, R, P)$, where $H_r$ is the height of a rectangle, $W_r$ is the width of the rectangle, $R$ is the ratio of $W_r$ to $H_r$ and, $P = H_r \times W_r$.

\item \emph{Symmetry Feature} $(S_H, S_V)$, where $S_H$ captures the horizontal symmetry and $S_V$ represents the vertical symmetry. We define $S_H$ and $S_V$ as follows: 
\begin{equation}
\left\{\begin{matrix}
S_H = \frac{2\sum\limits_{y=t}^{t+H_r}\sum\limits_{x=l}^{l+\frac{W_r}{2}}|I_H(x,y)-I_H(2l+W_r-x, y)|}{H_rW_r}\\
S_V = \frac{2\sum\limits_{x=l}^{l+W_r}\sum\limits_{y=t}^{t+\frac{H_r}{2}}|I_H(x,y)-I_H(x, 2t+H_r-y)|}{H_rW_r} 
\end{matrix}\right.
\label{eq:symetry}
\end{equation}
where, $(l, t)$ denotes the top-left corner point of the rectangle. Inclusion of this feature in our compound feature is motivated by the natural symmetry of human heads.
\item \emph{Zero Pixel Feature} $(N_0, R_0)$, where $N_0$ denotes the number of zero pixels appearing in the image area bounded by a rectangle, and $R_0 = \frac{N_0}{H_rW_r}$ is the rate of zero pixel appearance.
\item \emph{Expansion Ratio Feature} $\Psi \in \mathbb R^5$ contains the ratios of the area of a rectangle in $\sset F_H$ to five different rectangles achieved by using different expansion thresholds $\delta_h$  in Algorithm~\ref{alg:expandLocalMaximumPoints}. 
By varying the values of the expansion threshold we can expect different rectangles resulting for different kinds of objects in the scene.
Therefore, the expansion ratio feature provides important clues about an object being a head or not. In our original algorithm, we let $\delta_h = 15$ to arrive at the set $\sset E_H$. To compute the expansion ratio feature, we select the values of $\delta_h$ from $\{20, 25, 30, 35, 40\}$ to generate five different rectangles corresponding to each element of $\sset F_H$ and calculate $\Psi$ for each element.  
\end{itemize}
We concatenate the above mentioned four geometric features into a vector in $\mathbb R^{13}$. Notice that, although we do consider varied areas of $I_H$ in the above mentioned features, the compound feature only accounts for the information that is local to individual rectangles.
In the real-world scenarios, the relative locations of the rectangles (that we suspect to contain human heads) can provide useful information about a bounded  object being a human head or not. Therefore, we further define NRDF to account for this additional information. For each rectangle in $\sset F_H$, we compute NRDF as a vector in the 3D-space that is directed towards the center of the rectangle from the center of its nearest rectangle in our current set of head proposals. This feature is further illustrated in Fig.~\ref{fig:nrcvFeatures}. The resulting NRDF~$\in\mathbb R^3$ is concatenated with the above mentioned feature vector to finally arrive at our compound feature vector in $\mathbb R^{16}$.

\subsection{Tracking and Counting}
\label{sec:track}
Using the compound features introduced above, we refine the head proposals in $\sset F_H$. Notice that, this set is computed for a single depth frame in our approach. To eventually count the people passing through a scene, we must also track the trajectory of individual heads (i.e. people) in a continuous video stream. For that purpose, we exploit $\sset F_H$ in maintaining a record of head trajectories in the incoming video stream. We count the number of people passing by the camera by counting the number of trajectories disappearing in our records. We use the direction of movement to determine if the person has entered or exited the bus. Concrete technical details of this procedure are provided below.

To track individuals in the scene, we maintain a set of trajectories $\mathcal T$ for the continuous video stream. The set is initialized as `empty' when the stream starts.
With each frame the set gets updated by adding, removing or updating its elements.
An element of this set is given by $\{ \sset F_H^i, P^i\}$, where `$i$' indicates the $i^{\text{th}}$  element, and $P^i$ is the probability of that element bounding a human head. This probability is available to us from the SVM classifier trained to arrive at the refined set $\sset F_H$. In the text to follow, we refer to an element of $\mathcal T$ as a \emph{node} for brevity.

To update nodes with each coming frame, we first match the potential nodes of the new frame with the current nodes in $\mathcal T$. To that end, we compute $\eta = ||(x_o - x_n), (y_o - y_n), (s_o - s_n)||_2$, where $(x_o, y_o)$ indicates the center of the rectangle represented by a node in $\mathcal T$, $(x_n, y_n)$ is the center of a rectangle in the new frame, and $s_o$ and $s_n$ are the seed values for the respective rectangles. We consider two nodes to be matched if $\eta < \delta_m$, where we empirically fix the value of $\delta_m$. 
If a new node does not match any existing node, it is added to $\mathcal T$ as a new element. If an existing node in $\mathcal T$ is not updated for $Q$ consecutive frames, we remove that node from our set. The removed node increments our count of a person passing by the camera. When a node is removed, we determine the direction of the movement performed by the individual (i.e. `enter' or `exit') by analyzing the centers of the first and the last rectangle for that node. The information on the centers of rectangles (and their seeds), number of updates for each node, and the time stamp of the last update for each node are maintained in our approach by book-keeping. 

Using the simple strategy explained above, we can track the trajectories of individual objects in the scene. However, tracking of `human-heads' in the above method completely relies on the accuracy of  $\sset F_H$. If a non-head object still eludes our refinement  process discussed in the preceding Section, the approach may count extra individuals in the scene. To circumvent this problem we exploit the observation that human-heads generally follow similar trajectories in path-ways, which can be differentiated from the trajectories of non-head objects.  
Thus, we train binary SVM classifiers (one each for `entering' and `exiting' directions) to identify a given trajectory in $\mathcal T$ as `head' or `non-head'. We propose another feature for training the classifiers that is formed by concatenating a) the mean and variance of all nodes involved in a trajectory, b) the total number of updates for the trajectory, c) velocity of the trajectory computed as the rate of change in the center locations of the bounding rectangles, and d) the difference between the maximum and the minimum seed values for the trajectory. We use the SVMs trained over these features to refine our final count of the people entering or exiting the bus doors/path-ways.

\section{Experiment} \label{sec:E}
We evaluate the proposed method using our proposed dataset, PCDS, that contains a large number of pedestrians entering/exiting bus doors imaged by a Kinect camera installed on top of the door. The dataset  provides the opportunity to thoroughly evaluate the major components of our approach individually as well as analyze its performance for the overall task of people counting.
We first analyze the efficacy of our background subtraction procedure and compare its performance with the popular MOG~\cite{Zivkovic2004} and KNN-based methods~\cite{Zivkovic2006}.
Then, we separately analyze the performance of our method for the tasks of human head identification, human head tracking and finally, people counting as a whole.



\subsection{Background Removal}
Background removal is a major task in many surveillance related problems. For our approach, reliable background subtraction is necessary for the success of subsequent processing of video frames. Therefore, we separately analyze the performance of our method for this task.
We use the popular Gaussian mixture-based background segmentation method (MOG)~\cite{Zivkovic2004}  and the K-nearest neighbors (KNN) based method~\cite{Zivkovic2006} to benchmark our technique. We note that other approaches for background subtraction also exist, however the selected baseline methods are chosen for their well-established effectiveness for the depth videos.   
We carefully optimized parameter values of the baseline methods on our dataset using cross validation. For the proposed method, we empirically chose $n_c = 150$ and $n_{2c} = 500$ in all our experiments.

\begin{figure}[t]
\centering
\includegraphics[width=3in]{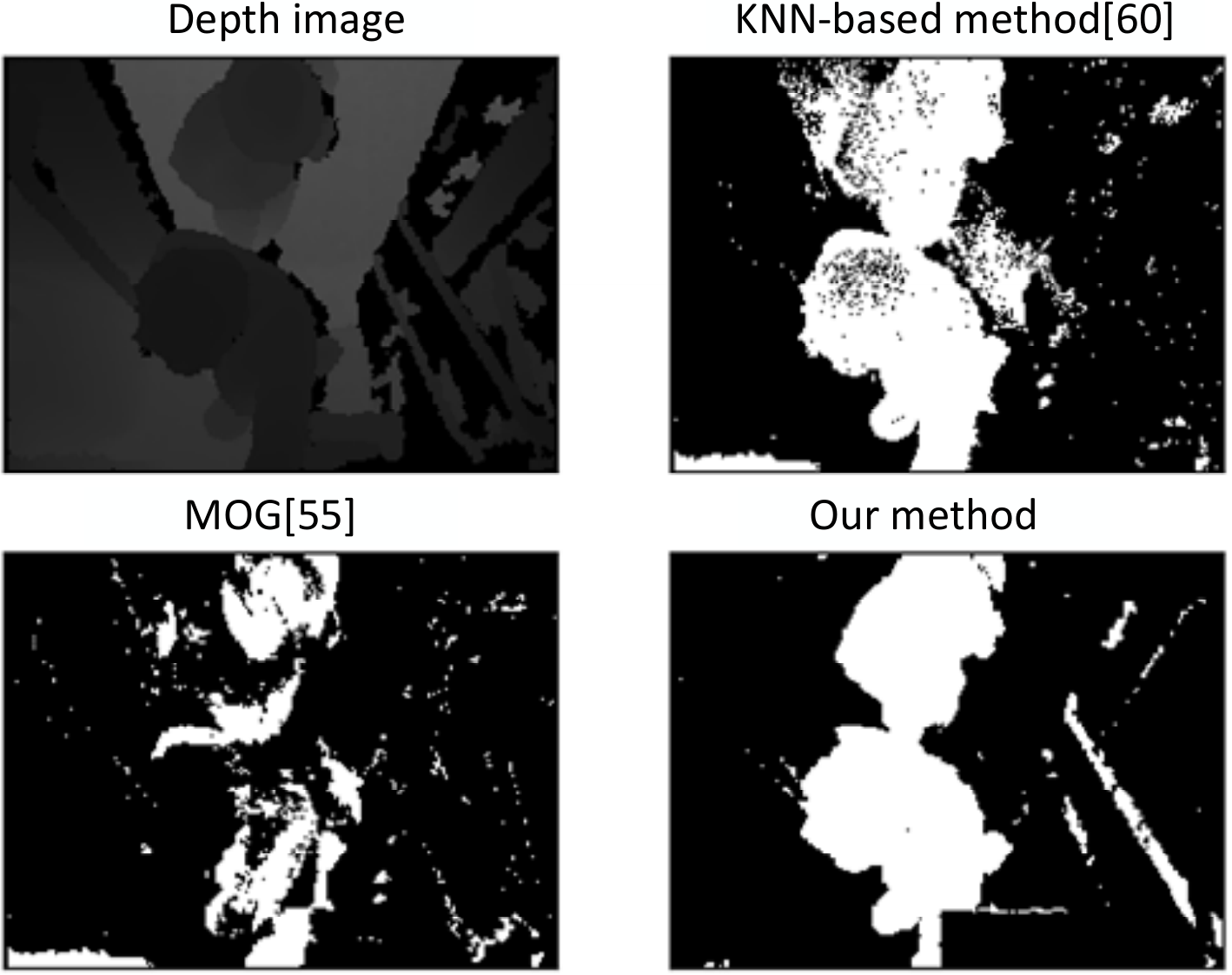}
\caption{Background subtract analysis. Top-left: Original depth frame. Top-right: Background identified by \cite{Zivkovic2006}. Bottom-left: Background identified by \cite{Zivkovic2004}. Bottom-left: Proposed method.}
\label{fig:backgroundSubtractComparison}
\end{figure}

Fig.~\ref{fig:backgroundSubtractComparison} shows a typical mask image generated by MOG~\cite{Zivkovic2004}, KNN-based method~\cite{Zivkovic2006},  and the proposed background subtraction procedure. As can be seen, the mask images generated by both KNN and MOG methods contain significant amount of noise which can be detrimental for the subsequent processing in our approach. On the other hand, the proposed method is able to preserve the masks of individual humans very well, with negligible noise. For further qualitative analysis of background subtraction, we also provide videos comparing our method with the existing approaches on the following URL: \url{https://youtu.be/oiuYq_Pfx6c}.





\begin{figure}[t]
\centering
\includegraphics[width=2in]{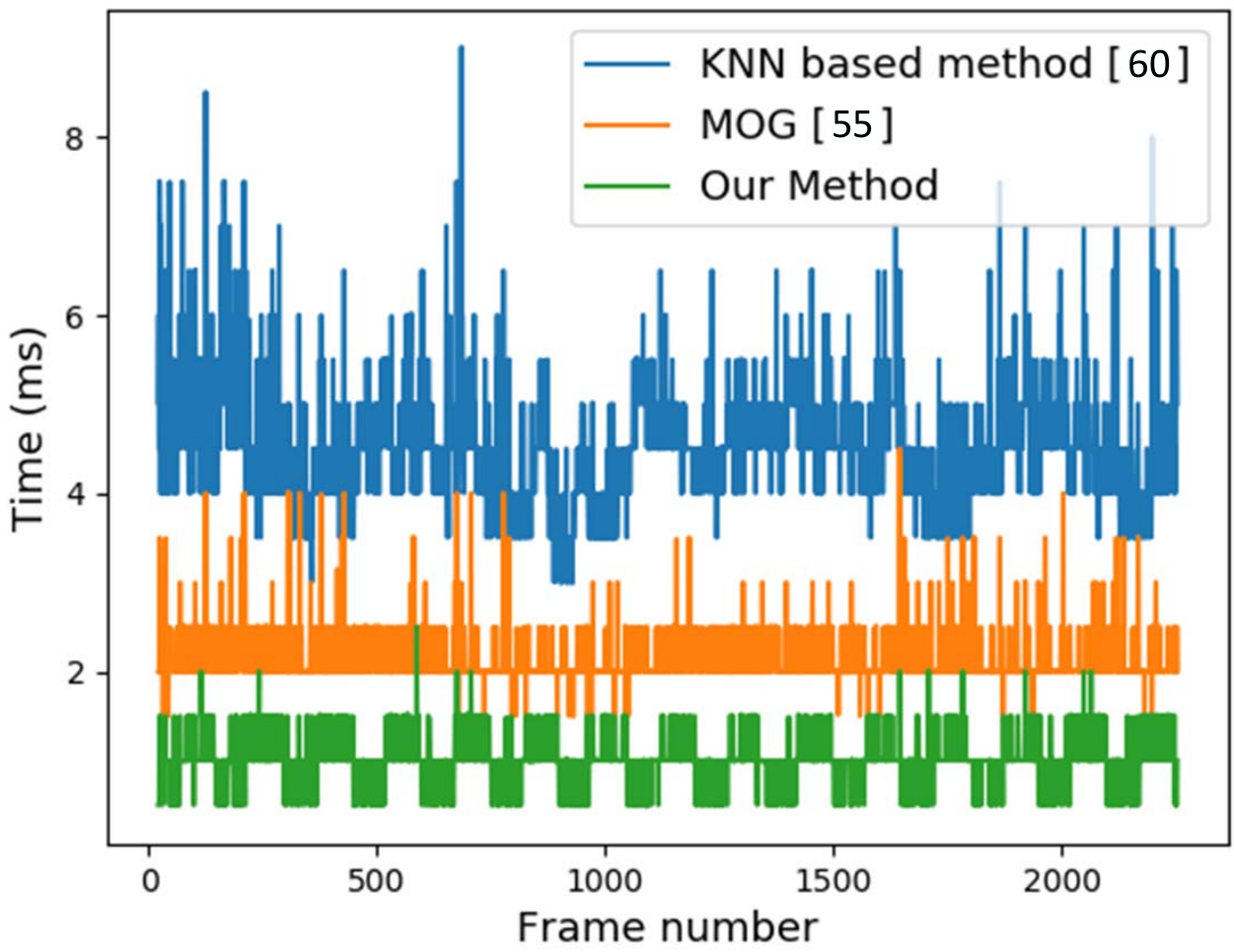}
\caption{Computational time for background subtraction for each frame. Timings for a sequence of 2,263 frames are shown.}
\label{fig:backgroundSubtractComparisonProcessingTime}
\end{figure}

Whereas our method achieves reliable background subtraction, it is also required to obtain those results efficiently for the overall task of \emph{real-time} people counting. We show the computational time (in ms) for processing each frame of a typical frame sequence in our dataset for the proposed approach and the baseline methods in  Fig~\ref{fig:backgroundSubtractComparisonProcessingTime}. The time is computed on a 1.7GHz processor with 2GB RAM for the task of background subtraction. The proposed method averages around 1.0 ms/frame in comparison to 2.1 ms/frame and 4.5 ms/frame of MOG and KNN-based method respectively. High quality background subtraction with a small time required to process each frame makes our background subtraction highly desirable for the broader problem of real-time people counting.

\subsection{Human Head Identification}
An essential component of counting people in our approach is to accurately identify human heads in the scene. We identify human heads by first generating candidate head proposals and then refining them. In our approach, the process of generating the candidate proposals is intentionally kept relaxed, and it also results in identifying multiple non-head objects in the scene (e.g. shoulders, bag-packs) to be considered as potential human heads. The refinement process (in Section~\ref{sec:HPR}) then discards the non-head objects to identify the human heads. 

\begin{figure}[t]
\centering
\includegraphics[width=3.5in]{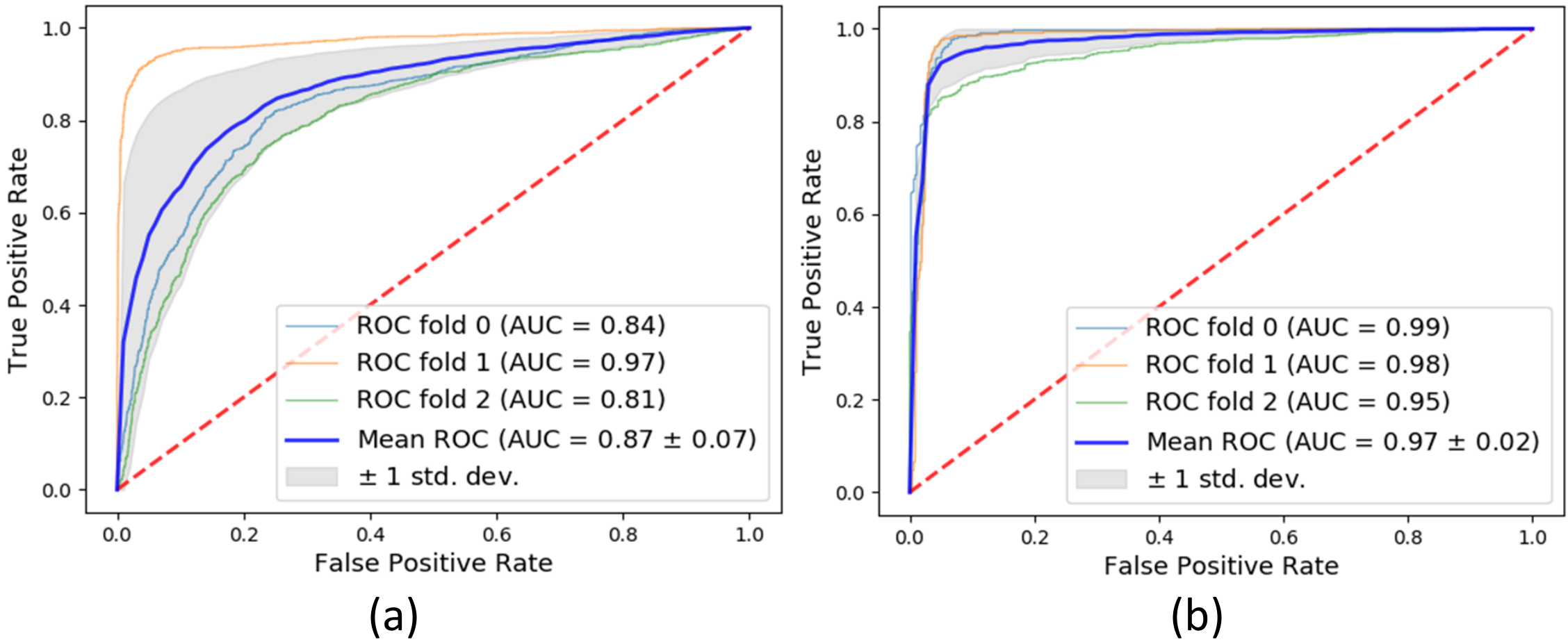}
\caption{The ROC curves of classifiers for head identification. a) People entering the buses: SVM parameters $\gamma=20.2$ and $C=16$. b) People exiting the buses: SVM parameters $\gamma=3.4$ and $C=520$.}
\label{fig:headClassifierAucCurve}
\end{figure}

To analyze the performance of our method for human head identification, we first manually labeled 12,148 rectangle proposals in height images for people entering the buses as `heads' and `non-heads'. These proposals were  generated automatically by the method in Section~\ref{sec:locationgHead}. We then trained and tested the SVM classifier employed in our approach using these proposals. We also performed the same routine for 10,108 candidate head proposals for the people exiting the buses.
The details of the train-test distributions and the labels of proposals used in this analysis are provided in Table~\ref{tab:labelnumber}.
In Fig.~\ref{fig:headClassifierAucCurve}, we show the ROC curves for the classifiers trained for the refinement of head proposals. The curves show  results of our three-fold experiments, with corresponding AUC values. 

From Fig.~\ref{fig:headClassifierAucCurve}, we can argue that the employed classifiers are able to identify human heads in the proposals successfully. We note that the classification performance depicted by Fig.~\ref{fig:headClassifierAucCurve} is better for the people exiting buses than for the people entering buses. The reason behind this phenomenon is that while providing the ground truth we only labeled those proposal rectangles as `heads' that bounded  complete human heads. For the case of people entering the buses, many half-heads appeared in the frames due to queuing of people on bus doors. On scrutiny, we found that most of those heads resulted in false positive identifications in our experiment. However, this is not problematic for the overall approach because the final results rely more strongly on tracking of heads on multiple frames, and the half-heads eventually transform into complete heads in the subsequent video frames. We also provide the details of precision, recall and the f1-scores for our head identification experiment in Table~\ref{tab:evalutionHeadTestSet}.

\begin{table}[t]
  \centering
  \caption{Summary of the labeled head proposals used}
    \begin{tabular}{ccccc}\hline
    \multicolumn{2}{c}{} & train & test & total \\\hline
    \multirow{2}[0]{*}{entering} & head  & 3123  & 2058  & 5181 \\
          & non-head & 4165  & 2802  & 6967 \\
    \multirow{2}[0]{*}{exiting} & head  & 3409  & 2172  & 5581 \\
          & non-head & 2872  & 1655  & 4527 \\\hline
    \end{tabular}%
  \label{tab:labelnumber}%
\end{table}%



\begin{table}[t]
  \centering
  \caption{Evaluation of head identification}
    \begin{tabular}{cccccc}\hline
    \multicolumn{2}{c}{} & precision & recall & f1-score & sample \\\hline
    \multirow{2}[0]{*}{entering} & head  & 0.92  & 0.92  & 0.92  & 2058 \\
          & non-head & 0.94  & 0.94  & 0.94  & 2802 \\
    \multirow{2}[0]{*}{exiting} & head  & 0.95  & 0.92  & 0.93  & 2172 \\
          & non-head & 0.97  & 0.98  & 0.98  & 1655 \\\hline
    \end{tabular}%
  \label{tab:evalutionHeadTestSet}%
\end{table}%



\subsection{Tracking}
Our overall approach relies strongly on the tracking method introduced in Section~\ref{sec:track}. Similar to the head identification method, we separately analyzed the tracking procedure by evaluating the performance of the classifier employed for tracking.
For that, we manually labeled 1,332 tracks in our dataset as `head' and `non-head' for people entering the buses. Among the labeled tracks, we used around 30\% samples for testing and the remaining samples were used for training the  classifier. We also followed the same routine for 1,330 tracks for  people exiting the buses. The information on the test-train distribution and the labels of the tracks used in our analysis is summarized in Table~\ref{tab:numberTracks}. We  empirically selected $\delta_m = 15$ and $Q =8$ in our experiments.



\begin{table}[t]
  \centering
  \caption{Summary of the labeled tracks used}
    \begin{tabular}{cccccc}\hline
    \multicolumn{2}{c}{} & train & test & total \\\hline
    \multirow{2}[0]{*}{entering} & head track & 442  & 183  & 625 \\
          & non-head track & 509  & 198  & 707 \\
    \multirow{2}[0]{*}{exiting} & head track & 544  & 226  & 770 \\
          & non-head track & 406  & 154  & 560 \\\hline
    \end{tabular}%
  \label{tab:numberTracks}%
\end{table}%

\begin{figure}[t]
\centering
\includegraphics[width=3.5in]{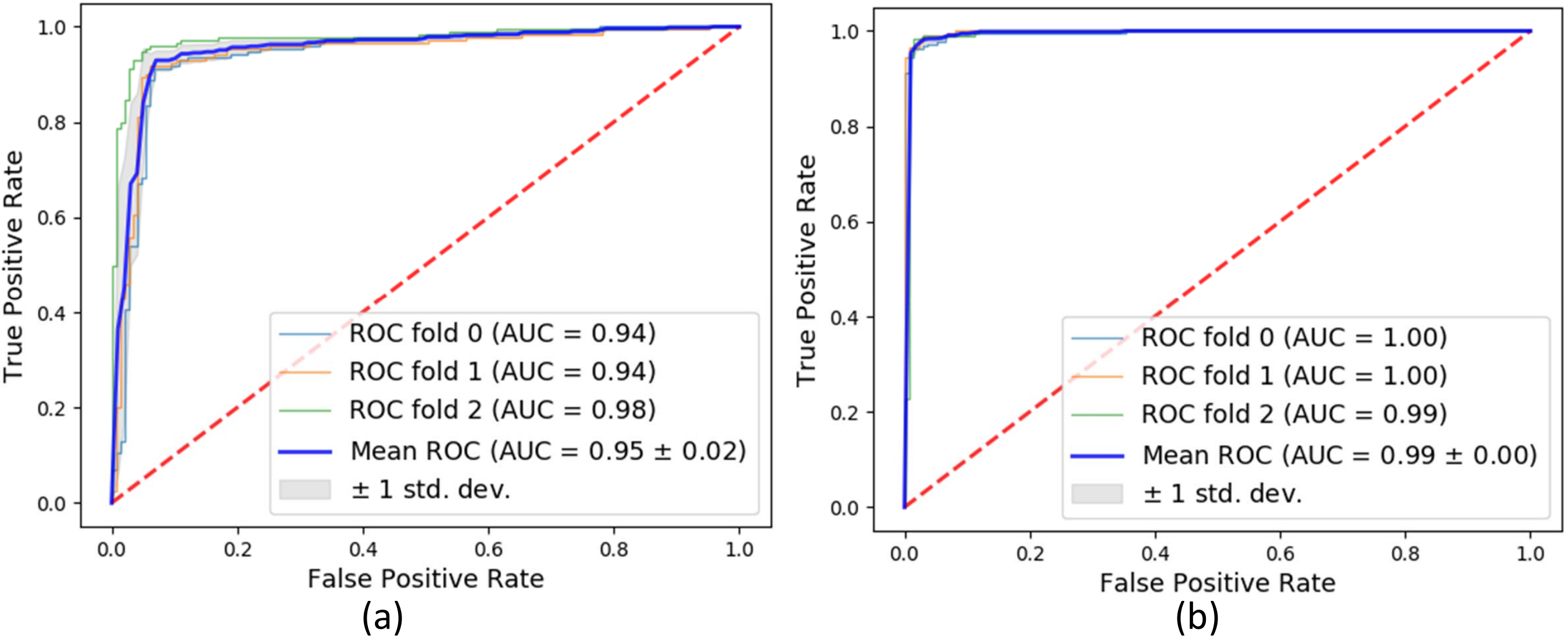}
\caption{The ROC curve of classifiers for tracking head trajectories. a) People entering the buses: SVM parameters  $\gamma=0.111$ and $C=368$. b) People exiting the buses: SVM parameters $\gamma=0.05882$ and $C=896$.}
\label{fig:trackClassifierAucCurve}
\end{figure}

In Fig.~\ref{fig:trackClassifierAucCurve}, we show the ROC curves of the classifiers used for head tracking in our approach. The figure also reports the AUC values for our three-fold experiments. It is easy to observe that our method is able to classify (i.e.~track) the trajectories of human heads very accurately for both `entering' and `exiting' scenarios.
Notice that no significant performance degradation is visible in Fig.~\ref{fig:trackClassifierAucCurve}a for the `entering' scenario, which was the case in Fig.~\ref{fig:headClassifierAucCurve}a. This is because tracking is performed over a sequence of frames and the incomplete heads (due to people queues) at the start of tracking eventually become irrelevant for the problem at hands. We also provide summary of the precision, recall and f1-scores of the tracking results in Table~\ref{tab:evaluationTrackClassifiers}.   The table indicates successful classification by the employed classifier.


\begin{table}[t]
  \centering
  \caption{Evaluation of tracking performance}
    \begin{tabular}{cccccc}\hline
    \multicolumn{2}{c}{} & precision & recall & f1-score & sample \\\hline
    \multirow{2}[0]{*}{entering} & head track & 0.92  & 0.97  & 0.94  & 183 \\
          & non-head track & 0.97  & 0.92  & 0.95  & 198 \\
    \multirow{2}[0]{*}{exiting} & head track & 0.98  & 0.97  & 0.98  & 226 \\
          & non-head track & 0.96  & 0.97  & 0.97  & 154 \\\hline
    \end{tabular}%
  \label{tab:evaluationTrackClassifiers}%
\end{table}%

\subsection{People Counting}
The main objective of our approach is to perform people counting in real-time. We evaluated the people counting performance of our approach using 2,000 test videos from PCDS. We used detection rate `$\Delta$' as the metric for evaluations, which is defined as follows.
\begin{equation}
\Delta = \frac{\sum\limits_{i=1}^{N_V}|n_{i}-\widetilde{n_{i}}|}{\sum\limits_{i=1}^{N_V}n_{i}},
\end{equation}
where $N_V$ is the total number of videos in the test data, $n_i$ is the number of people passing in the $i^{\text{th}}$ video, and $\widetilde{n_i}$ denotes the estimated number of people in the $i^{\text{th}}$ video.



\begin{table}[t]
  \centering
  \caption{People counting accuracy on $\rm{PCDS}$}
    \begin{tabular}{rcccc}\hline
                  &    $N^-C^-$    &    $N^-C^+$       & $N^+C^-$    &     $N^+C^+$       \\\hline
    entering     & 85.40\%         &     83.25\%     & 77.54\%     & 75.32\% \\
    exiting      & 93.04\%         &     92.66\%     & 93.71\%     & 91.30\% \\ \hline
    \end{tabular}%
  \label{tab:detectionRate}%
\end{table}%

\begin{figure}[t]
\centering
\includegraphics[width=2in]{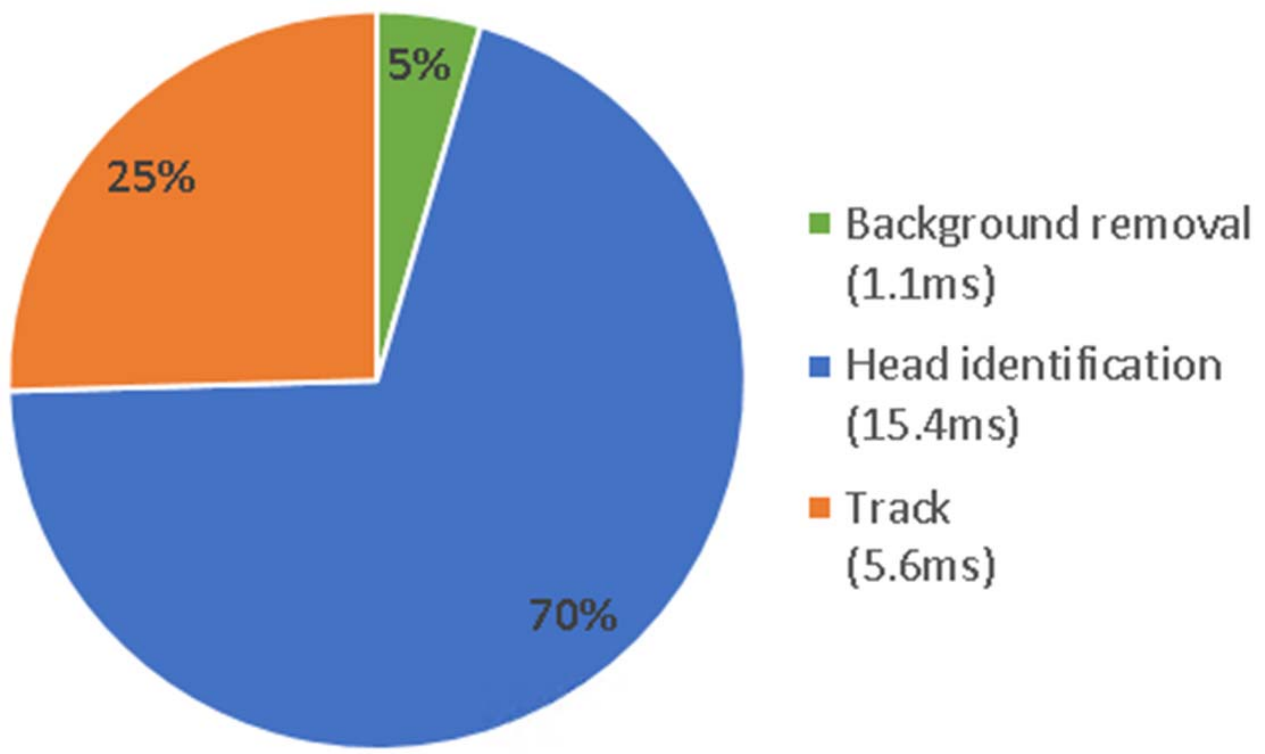}
\caption{Distribution of average computational time of our method. A single frame required 22.1ms on average, which results in processing approximately 45 frame per second on a 1.7GHz Intel processor with 2GB RAM.}
\label{fig:run_time}
\end{figure}

In Table~\ref{tab:detectionRate}, we report the detection rates of our method for the four different categories of videos introduced in Section~\ref{sec:PCDS_Tax}. We separately report the results for the `entering' and `exiting' scenarios.
Based on these results, we can argue that the performance of our approach is acceptable for both scenarios.
In PCDS, most of the people entering the buses use  the front door. It was observed that due to significant glare from the glass of the front doors, the videos often contained large amount of noise. This generally made  people counting at the front doors in the dataset more challenging. 
Nevertheless, the approach shows reasonable overall performance given the practical real-world conditions of the dataset.


Considering the potential low on-board computational capacity available for our method in the real-world deployment, we used a less powerful 1.7GHz Intel processor with 2GB RAM for evaluating our approach. Figure \ref{fig:run_time} shows the distribution of the  average computational time taken by different components of our method on the used processor for a single frame.
The overall average time for processing a single frame is about 22.1ms, amounting to approximately 45 frames per second, which can be considered as real-time performance. 

\section{Conclusion} \label{sec:C}
This article makes two important contributions to the problem of `people counting' in real-world scenarios. Firstly, it presents the first large-scale benchmark public dataset for the problem. This dataset contains recorded depth videos, color videos and CSV format files with the labels containing the number of people passing through different scenes of bus doors.
The videos account for a large variability in scene illumination, clutter, noise and other factors in the real-world environment, which makes the dataset particularly challenging. Secondly, the article presents a method for real-time people counting in cluttered scenes and evaluates the performance on the proposed dataset. 
The proposed method utilizes the depth video stream and computes a normalized height image of the scene after removing the background. 
The height image is essentially a projection of the scene depth directly below the camera, which helps in a clear segmentation of individual objects in the scene.
This projection is used to identify heads of individuals in the scene. We utilize a 3D human model and adapt a seed fill method to reliably detect  human heads. We also propose a compound feature for height images, that is utilized in our approach for head identification.
Once reliably detected, individual human heads are tracked to compute their trajectory which is eventually utilized for people counting.
We ascertain the effectiveness of our method by applying it to the proposed dataset. Our benchmark dataset will play a major role in advancing research in the areas of RGB-video, Depth-video and RGBD-video based people counting.

\ifCLASSOPTIONcaptionsoff
  \newpage
\fi



\bibliographystyle{IEEEtran}
\bibliography{peopleCounting}

\begin{thebibliography}{10}
\providecommand{\url}[1]{#1}
\csname url@samestyle\endcsname
\providecommand{\newblock}{\relax}
\providecommand{\bibinfo}[2]{#2}
\providecommand{\BIBentrySTDinterwordspacing}{\spaceskip=0pt\relax}
\providecommand{\BIBentryALTinterwordstretchfactor}{4}
\providecommand{\BIBentryALTinterwordspacing}{\spaceskip=\fontdimen2\font plus
\BIBentryALTinterwordstretchfactor\fontdimen3\font minus
  \fontdimen4\font\relax}
\providecommand{\BIBforeignlanguage}[2]{{%
\expandafter\ifx\csname l@#1\endcsname\relax
\typeout{** WARNING: IEEEtran.bst: No hyphenation pattern has been}%
\typeout{** loaded for the language `#1'. Using the pattern for}%
\typeout{** the default language instead.}%
\else
\language=\csname l@#1\endcsname
\fi
#2}}
\providecommand{\BIBdecl}{\relax}
\BIBdecl

\bibitem{6817563}
B.~Barabino, M.~D. Francesco, and S.~Mozzoni, ``An offline framework for
  handling automatic passenger counting raw data,'' \emph{IEEE Transactions on
  Intelligent Transportation Systems}, vol.~15, no.~6, pp. 2443--2456, Dec
  2014.

\bibitem{7837703}
Y.~Wang, D.~Zhang, L.~Hu, Y.~Yang, and L.~H. Lee, ``A data-driven and optimal
  bus scheduling model with time-dependent traffic and demand,'' \emph{IEEE
  Transactions on Intelligent Transportation Systems}, vol.~18, no.~9, pp.
  2443--2452, Sept 2017.

\bibitem{ceder1984bus}
A.~Ceder, ``Bus frequency determination using passenger count data,''
  \emph{Transportation Research Part A: General}, vol.~18, no. 5-6, pp.
  439--453, 1984.

\bibitem{Zhang2012c}
Z.~Zhang, ``{Microsoft kinect sensor and its effect},'' \emph{IEEE Multimedia},
  vol.~19, no.~2, pp. 4--10, 2012.

\bibitem{Freedman2010}
B.~Freedman, S.~Alexander, M.~Machline, and Y.~Arieli, ``{Depth Mapping Using
  Projected Patterns},'' \emph{WO Patent WO2008}, vol.~1, no.~19, p.~12, 2010.

\bibitem{Mallick2014}
T.~Mallick, P.~P. Das, and A.~K. Majumdar, ``{Characterizations of noise in
  Kinect depth images: A review},'' pp. 1731--1740, 2014.

\bibitem{Barandiaran2008a}
J.~Barandiaran, B.~Murguia, and F.~Boto, ``{Real-time people counting using
  multiple lines},'' in \emph{WIAMIS 2008 - Proceedings of the 9th
  International Workshop on Image Analysis for Multimedia Interactive
  Services}, 2008, pp. 159--162.

\bibitem{Fradi2012}
H.~Fradi and J.~L. Dugelay, ``{Low level crowd analysis using frame-wise
  normalized feature for people counting},'' in \emph{WIFS 2012 - Proceedings
  of the 2012 IEEE International Workshop on Information Forensics and
  Security}, 2012, pp. 246--251.

\bibitem{Antonini2006}
G.~Antonini and J.~P. Thiran, ``{Counting pedestrians in video sequences using
  trajectory clustering},'' \emph{IEEE Transactions on Circuits and Systems for
  Video Technology}, vol.~16, no.~8, pp. 1008--1020, 2006.

\bibitem{Topkaya2014}
I.~S. Topkaya, H.~Erdogan, and F.~Porikli, ``{Counting people by clustering
  person detector outputs},'' in \emph{11th IEEE International Conference on
  Advanced Video and Signal-Based Surveillance, AVSS 2014}, 2014, pp. 313--318.

\bibitem{Zeng2010}
C.~Zeng and H.~Ma, ``{Robust head-shoulder detection by PCA-based multilevel
  HOG-LBP detector for people counting},'' in \emph{Proceedings - International
  Conference on Pattern Recognition}, 2010, pp. 2069--2072.

\bibitem{DelPizzo2016a}
\BIBentryALTinterwordspacing
L.~Del~Pizzo, P.~Foggia, A.~Greco, G.~Percannella, and M.~Vento, ``{Counting
  people by RGB or depth overhead cameras},'' \emph{Pattern Recognition
  Letters}, vol.~81, pp. 41--50, 2016. [Online]. Available:
  \url{http://dx.doi.org/10.1016/j.patrec.2016.05.033}
\BIBentrySTDinterwordspacing

\bibitem{Chen2012a}
\BIBentryALTinterwordspacing
C.~Chen, T.~Chen, D.~Wang, and T.~Chen, ``{A cost-effective people-counter for
  a crowd of moving people based on two-stage segmentation},'' \emph{J Inform
  Hiding Multimedia {\ldots}}, vol.~3, no.~1, pp. 12--23, 2012. [Online].
  Available:
  \url{http://www.jihmsp.org/~jihmsp/2012/vol3/JIH-MSP-2012-01-002.pdf}
\BIBentrySTDinterwordspacing

\bibitem{Li2017}
G.~Li, P.~Ren, X.~Lyu, and H.~Zhang, ``{Real-Time Top-View People Counting
  Based on a Kinect and NVIDIA Jetson TK1 Integrated Platform},'' \emph{IEEE
  International Conference on Data Mining Workshops, ICDMW}, pp. 468--473,
  2017.

\bibitem{Gao2016}
\BIBentryALTinterwordspacing
C.~Gao, J.~Liu, Q.~Feng, and J.~Lv, ``{People-flow counting in complex
  environments by combining depth and color information},'' \emph{Multimedia
  Tools and Applications}, vol.~75, no.~15, pp. 9315--9331, 8 2016. [Online].
  Available: \url{http://link.springer.com/10.1007/s11042-016-3344-z}
\BIBentrySTDinterwordspacing

\bibitem{Liu2017}
G.~Liu, Z.~Yin, Y.~Jia, and Y.~Xie, ``{Passenger flow estimation based on
  convolutional neural network in public transportation system},''
  \emph{Knowledge-Based Systems}, vol. 123, pp. 102--115, 2017.

\bibitem{Kocak2017}
\BIBentryALTinterwordspacing
Y.~P. Kocak and S.~Sevgen, ``{Detecting and counting people using real-time
  directional algorithms implemented by compute unified device architecture},''
  \emph{Neurocomputing}, vol. 248, pp. 105--111, 2017. [Online]. Available:
  \url{http://dx.doi.org/10.1016/j.neucom.2016.08.137}
\BIBentrySTDinterwordspacing

\bibitem{Chen2014}
K.~Chen and J.~K. K{\"{a}}m{\"{a}}r{\"{a}}inen, ``{Learning to count with
  back-propagated information},'' in \emph{Proceedings - International
  Conference on Pattern Recognition}, 2014, pp. 4672--4677.

\bibitem{Chan2012a}
A.~B. Chan and N.~Vasconcelos, ``{Counting people with low-level features and
  bayesian regression},'' \emph{IEEE Transactions on Image Processing},
  vol.~21, no.~4, pp. 2160--2177, 2012.

\bibitem{Zhang2015}
C.~Zhang, H.~Li, X.~Wang, and X.~Yang, ``{Cross-scene crowd counting via deep
  convolutional neural networks},'' in \emph{Proceedings of the IEEE Computer
  Society Conference on Computer Vision and Pattern Recognition}, vol.
  07-12-June, 2015, pp. 833--841.

\bibitem{Idrees2013}
H.~Idrees, I.~Saleemi, C.~Seibert, and M.~Shah, ``{Multi-source multi-scale
  counting in extremely dense crowd images},'' in \emph{Proceedings of the IEEE
  Computer Society Conference on Computer Vision and Pattern Recognition},
  2013, pp. 2547--2554.

\bibitem{Cong2009}
Y.~Cong, H.~Gong, S.~C. Zhu, and Y.~Tang, ``{Flow mosaicking: Real-time
  pedestrian counting without Scene-specific learning},'' in \emph{2009 IEEE
  Computer Society Conference on Computer Vision and Pattern Recognition
  Workshops, CVPR Workshops 2009}, 2009, pp. 1093--1100.

\bibitem{Benabbas2010a}
Y.~Benabbas, N.~Ihaddadene, T.~Yahiaoui, T.~Urruty, and C.~Djeraba,
  ``{Spatio-temporal optical flow analysis for people counting},'' in
  \emph{Proceedings - IEEE International Conference on Advanced Video and
  Signal Based Surveillance, AVSS 2010}, 2010, pp. 212--217.

\bibitem{Cong2009a}
Y.~Cong, H.~Gong, S.~C. Zhu, and Y.~Tang, ``{Flow mosaicking: Real-time
  pedestrian counting without Scene-specific learning},'' in \emph{2009 IEEE
  Computer Society Conference on Computer Vision and Pattern Recognition
  Workshops, CVPR Workshops 2009}, 2009, pp. 1093--1100.

\bibitem{Neal2000}
R.~M. Neal, ``{Markov Chain Sampling Methods for Dirichlet Process Mixture
  Models},'' \emph{Journal of Computational and Graphical Statistics}, vol.~9,
  no.~2, pp. 249--265, 2000.

\bibitem{Brostow2006a}
G.~J. Brostow and R.~Cipolla, ``{Unsupervised bayesian detection of independent
  motion in crowds},'' in \emph{Proceedings of the IEEE Computer Society
  Conference on Computer Vision and Pattern Recognition}, vol.~1, 2006, pp.
  594--601.

\bibitem{Rabaud2006}
V.~Rabaud and S.~Belongie, ``{Counting crowded moving objects},'' in
  \emph{Proceedings of the IEEE Computer Society Conference on Computer Vision
  and Pattern Recognition}, vol.~1, 2006, pp. 705--711.

\bibitem{Lucas1981}
\BIBentryALTinterwordspacing
B.~D. Lucas and T.~Kanade, ``{An Iterative Image Registration Technique with an
  Application to Stereo Vision},'' \emph{Imaging}, vol. 130, no.~x, pp.
  674--679, 1981. [Online]. Available:
  \url{http://citeseerx.ist.psu.edu/viewdoc/download?doi=10.1.1.49.2019&amp;rep=rep1&amp;type=pdf}
\BIBentrySTDinterwordspacing

\bibitem{Dalal2005a}
N.~Dalal and B.~Triggs, ``{Histograms of oriented gradients for human
  detection},'' in \emph{Proceedings - 2005 IEEE Computer Society Conference on
  Computer Vision and Pattern Recognition, CVPR 2005}, vol.~I, 2005, pp.
  886--893.

\bibitem{Ojala2002}
T.~Ojala, M.~Pietik{\"{a}}inen, and T.~M{\"{a}}enp{\"{a}}{\"{a}},
  ``{Multiresolution gray-scale and rotation invariant texture classification
  with local binary patterns},'' \emph{IEEE Transactions on Pattern Analysis
  and Machine Intelligence}, vol.~24, no.~7, pp. 971--987, 2002.

\bibitem{Wold1987}
S.~Wold, K.~Esbensen, and P.~Geladi, ``{Principal component analysis},''
  \emph{Chemometrics and Intelligent Laboratory Systems}, vol.~2, no. 1-3, pp.
  37--52, 1987.

\bibitem{Antic2009}
B.~Anti{\'{c}}, D.~Leti{\'{c}}, D.~{\'{C}}ulibrk, and V.~Crnojevi{\'{c}},
  ``{K-means based segmentation for real-time zenithal people counting},'' in
  \emph{Proceedings - International Conference on Image Processing, ICIP},
  2009, pp. 2565--2568.

\bibitem{Garcia2013}
J.~Garcia, A.~Gardel, I.~Bravo, J.~L. Lazaro, M.~Martinez, and D.~Rodriguez,
  ``{Directional people counter based on head tracking},'' \emph{IEEE
  Transactions on Industrial Electronics}, vol.~60, no.~9, pp. 3991--4000,
  2013.

\bibitem{Julier1997}
\BIBentryALTinterwordspacing
S.~J. Julier and J.~K. Uhlmann, ``{New extension of the Kalman filter to
  nonlinear systems},'' \emph{Int Symp AerospaceDefense Sensing Simul and
  Controls}, vol.~3, p. 182, 1997. [Online]. Available:
  \url{http://link.aip.org/link/?PSI/3068/182/1&Agg=doi%5Cnhttp://proceedings.spiedigitallibrary.org/proceeding.aspx?doi=10.1117/12.280797}
\BIBentrySTDinterwordspacing

\bibitem{Chaohui2007}
Z.~C.~Z. Chaohui, D.~X.~D. Xiaohui, X.~S.~X. Shuoyu, S.~Z.~S. Zheng, and
  L.~M.~L. Min, ``{An Improved Moving Object Detection Algorithm Based on Frame
  Difference and Edge Detection},'' \emph{Fourth International Conference on
  Image and Graphics (ICIG 2007)}, pp. 519--523, 2007.

\bibitem{Kurilkin2016}
A.~V. Kurilkin and S.~V. Ivanov, ``{A Comparison of Methods to Detect People
  Flow Using Video Processing},'' in \emph{Procedia Computer Science}, vol.
  101, 2016, pp. 125--134.

\bibitem{Terada1999}
K.~Terada, D.~Yoshida, S.~Oe, and J.~Yamaguchi, ``{A method of counting the
  passing people by using the stereo images},'' \emph{Proceedings 1999
  International Conference on Image Processing (Cat. 99CH36348)}, vol.~2, pp.
  338--342, 1999.

\bibitem{Kristoffersen2016a}
M.~S. Kristoffersen, J.~V. Dueholm, R.~Gade, and T.~B. Moeslund, ``{Pedestrian
  counting with occlusion handling using stereo thermal cameras},''
  \emph{Sensors (Switzerland)}, vol.~16, no.~1, 2016.

\bibitem{Zhang2012b}
Z.~Zhang, ``{Microsoft kinect sensor and its effect},'' pp. 4--10, 2012.

\bibitem{Zhang2012a}
\BIBentryALTinterwordspacing
X.~Zhang, J.~Yan, S.~Feng, Z.~Lei, D.~Yi, and S.~Z. Li, ``{Water filling:
  Unsupervised people counting via vertical kinect sensor},'' in
  \emph{Proceedings - 2012 IEEE 9th International Conference on Advanced Video
  and Signal-Based Surveillance, AVSS 2012}.\hskip 1em plus 0.5em minus
  0.4em\relax IEEE, 9 2012, pp. 215--220. [Online]. Available:
  \url{http://ieeexplore.ieee.org/document/6328019/}
\BIBentrySTDinterwordspacing

\bibitem{Pizzo2015}
L.~D. Pizzo, P.~Foggia, A.~Greco, G.~Percannella, and M.~Vento, ``{A versatile
  and effective method for counting people on either RGB or depth overhead
  cameras},'' in \emph{2015 IEEE International Conference on Multimedia Expo
  Workshops (ICMEW)}, 2015, pp. 1--6.

\bibitem{Rauter2013}
M.~Rauter, ``{Reliable human detection and tracking in top-view depth
  images},'' in \emph{IEEE Computer Society Conference on Computer Vision and
  Pattern Recognition Workshops}, 2013, pp. 529--534.

\bibitem{Vera2016a}
\BIBentryALTinterwordspacing
P.~Vera, S.~Monjaraz, and J.~Salas, ``{Counting pedestrians with a zenithal
  arrangement of depth cameras},'' \emph{Machine Vision and Applications},
  vol.~27, no.~2, pp. 303--315, 2 2016. [Online]. Available:
  \url{http://link.springer.com/10.1007/s00138-015-0739-1}
\BIBentrySTDinterwordspacing

\bibitem{najman1996geodesic}
L.~Najman and M.~Schmitt, ``{Geodesic saliency of watershed contours and
  hierarchical segmentation},'' \emph{IEEE Transactions on pattern analysis and
  machine intelligence}, vol.~18, no.~12, pp. 1163--1173, 1996.

\bibitem{kuhn1955hungarian}
H.~W. Kuhn, ``{The Hungarian method for the assignment problem},'' \emph{Naval
  Research Logistics (NRL)}, vol.~2, no. 1-2, pp. 83--97, 1955.

\bibitem{Ukidave2015}
Y.~Ukidave, D.~Kaeli, U.~Gupta, and K.~Keville., ``{Performance of the NVIDIA
  Jetson TK1 in HPC},'' in \emph{Proceedings - IEEE International Conference on
  Cluster Computing, ICCC}, vol. 2015-Octob, 2015, pp. 533--534.

\bibitem{Bondi2014}
E.~Bondi, L.~Seidenari, A.~D. Bagdanov, and A.~Del~Bimbo, ``{Real-time people
  counting from depth imagery of crowded environments},'' \emph{11th IEEE
  International Conference on Advanced Video and Signal-Based Surveillance,
  AVSS 2014}, pp. 337--342, 2014.

\bibitem{Liu2013}
\BIBentryALTinterwordspacing
J.~Liu, Y.~Liu, Y.~Cui, and Y.~Chen, ``{Real-time human detection and tracking
  in complex environments using single RGBD camera.}'' \emph{Icip}, pp.
  3088--3092, 9 2013. [Online]. Available:
  \url{http://ieeexplore.ieee.org/document/6738636/
  http://2013.ieeeicip.org/proc/pdfs/0003088.pdf}
\BIBentrySTDinterwordspacing

\bibitem{Zhang2016}
\BIBentryALTinterwordspacing
G.~Zhang, J.~Liu, L.~Tian, and Y.~Q. Chen, ``{Reliably detecting humans with
  RGB-D camera with physical blob detector followed by learning-based
  filtering},'' in \emph{ICASSP, IEEE International Conference on Acoustics,
  Speech and Signal Processing - Proceedings}, vol. 2016-May.\hskip 1em plus
  0.5em minus 0.4em\relax IEEE, 3 2016, pp. 2004--2008. [Online]. Available:
  \url{http://ieeexplore.ieee.org/document/7472028/}
\BIBentrySTDinterwordspacing

\bibitem{jia2014caffe}
Y.~Jia, E.~Shelhamer, J.~Donahue, S.~Karayev, J.~Long, R.~Girshick,
  S.~Guadarrama, and T.~Darrell, ``{Caffe: Convolutional architecture for fast
  feature embedding},'' in \emph{Proceedings of the 22nd ACM international
  conference on Multimedia}.\hskip 1em plus 0.5em minus 0.4em\relax ACM, 2014,
  pp. 675--678.

\bibitem{LeCun2015}
\BIBentryALTinterwordspacing
Y.~LeCun, Y.~Bengio, and G.~Hinton, ``{Deep learning},'' \emph{Nature}, vol.
  521, no. 7553, pp. 436--444, 2015. [Online]. Available:
  \url{http://www.nature.com/doifinder/10.1038/nature14539}
\BIBentrySTDinterwordspacing

\bibitem{Abadi2016}
\BIBentryALTinterwordspacing
M.~Abadi, P.~Barham, J.~Chen, Z.~Chen, A.~Davis, J.~Dean, M.~Devin,
  S.~Ghemawat, G.~Irving, M.~Isard, M.~Kudlur, J.~Levenberg, R.~Monga,
  S.~Moore, D.~G. Murray, B.~Steiner, P.~Tucker, V.~Vasudevan, P.~Warden,
  M.~Wicke, Y.~Yu, X.~Zheng, and G.~Brain, ``{TensorFlow: A System for
  Large-Scale Machine Learning TensorFlow: A system for large-scale machine
  learning},'' in \emph{12th USENIX Symposium on Operating Systems Design and
  Implementation (OSDI '16)}, 2016, pp. 265--284. [Online]. Available:
  \url{https://www.usenix.org/conference/osdi16/technical-sessions/presentation/abadi}
\BIBentrySTDinterwordspacing

\bibitem{Cao2017}
J.~Cao, Y.~Pang, and X.~Li, ``{Learning multilayer channel features for
  pedestrian detection},'' \emph{IEEE Transactions on Image Processing},
  vol.~26, no.~7, pp. 3210--3220, 2017.

\bibitem{Ciregan2012a}
\BIBentryALTinterwordspacing
D.~Ciregan, U.~Meier, and J.~Schmidhuber, ``{Multi-column deep neural networks
  for image classification},'' in \emph{Computer Vision and Pattern Recognition
  (CVPR)}, 2012, pp. 3642--3649. [Online]. Available:
  \url{http://ieeexplore.ieee.org/xpls/abs_all.jsp?arnumber=6248110}
\BIBentrySTDinterwordspacing

\bibitem{zhang2014fast}
K.~Zhang, L.~Zhang, Q.~Liu, D.~Zhang, and M.-H. Yang, ``{Fast visual tracking
  via dense spatio-temporal context learning},'' in \emph{European Conference
  on Computer Vision}.\hskip 1em plus 0.5em minus 0.4em\relax Springer, 2014,
  pp. 127--141.

\bibitem{Wei2017}
\BIBentryALTinterwordspacing
X.~Wei, J.~Du, M.~Liang, and L.~Ye, ``{Boosting Deep Attribute Learning via
  Support Vector Regression for Fast Moving Crowd Counting},'' \emph{Pattern
  Recognition Letters}, vol.~47, pp. 178--193, 2017. [Online]. Available:
  \url{http://linkinghub.elsevier.com/retrieve/pii/S0167865517304415}
\BIBentrySTDinterwordspacing

\bibitem{Simonyan2014}
\BIBentryALTinterwordspacing
K.~Simonyan and A.~Zisserman, ``{Very Deep Convolutional Networks for
  Large-Scale Image Recognition},'' \emph{arXiv preprint arXiv:1409.1556}, pp.
  1--13, 2014. [Online]. Available: \url{http://arxiv.org/abs/1409.1556}
\BIBentrySTDinterwordspacing

\bibitem{Kaewtrakulpong2001}
\BIBentryALTinterwordspacing
P.~Kaewtrakulpong and R.~Bowden, ``{An Improved Adaptive Background Mixture
  Model for Real- time Tracking with Shadow Detection},'' \emph{Advanced Video
  Based Surveillance Systems}, pp. 1--5, 2001. [Online]. Available:
  \url{http://personal.ee.surrey.ac.uk/Personal/R.Bowden/publications/avbs01/avbs01.pdf}
\BIBentrySTDinterwordspacing

\bibitem{Zivkovic2004}
\BIBentryALTinterwordspacing
Z.~Zivkovic, ``{Improved adaptive Gaussian mixture model for background
  subtraction},'' in \emph{Proceedings of the 17th International Conference on
  Pattern Recognition}, vol.~2, no.~2, 2004, pp. 28--31. [Online]. Available:
  \url{http://ieeexplore.ieee.org/document/1333992/}
\BIBentrySTDinterwordspacing

\bibitem{Del-Blanco2014}
C.~R. del Blanco, T.~Mantec{\'{o}}n, M.~Camplani, F.~Jaureguizar, L.~Salgado,
  and N.~Garc{\'{i}}a, ``{Foreground segmentation in depth imagery using depth
  and spatial dynamic models for video surveillance applications},''
  \emph{Sensors (Switzerland)}, vol.~14, no.~2, pp. 1961--1987, 2014.

\bibitem{Chacon-Murguia2016}
M.~I. Chacon-Murguia, O.~A. Chavez-Montes, and J.~A. Ramirez-Quintana,
  ``{Background modeling on depth video sequences using self-organizing
  retinotopic maps},'' in \emph{Proceedings of the International Joint
  Conference on Neural Networks}, vol. 2016-Octob, 2016, pp. 1090--1095.

\bibitem{Bjorck1996}
A.~Bjorck, \emph{{Numerical Methods for Least Squares Problems}}.\hskip 1em
  plus 0.5em minus 0.4em\relax Siam, 1996.

\bibitem{torbert2016applied}
S.~Torbert, \emph{{Applied computer science}}.\hskip 1em plus 0.5em minus
  0.4em\relax Springer, 2016.

\bibitem{Zivkovic2006}
Z.~Zivkovic and F.~Van Der~Heijden, ``{Efficient adaptive density estimation
  per image pixel for the task of background subtraction},'' \emph{Pattern
  Recognition Letters}, vol.~27, no.~7, pp. 773--780, 2006.

\end{thebibliography}

\begin{IEEEbiography}[{\includegraphics[width=1in,height=1.25in,clip,keepaspectratio]{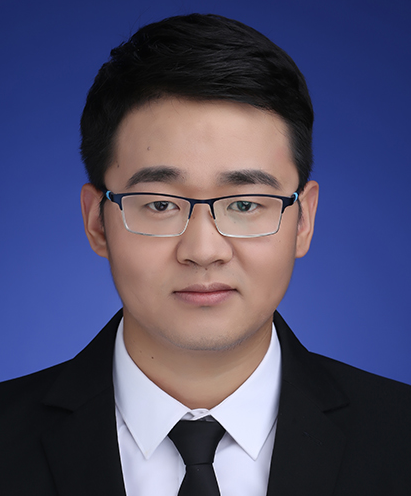}}]{ShiJie Sun} received his B.S. in software engineering from The University of Chang'an University and is currently working towards the Ph.D. degree in intelligent transportation and information system engineering with Chang'an University. Currently, he is a visiting (joint) Ph.D. candidate at the University of Western Australia since October 2017. His research interests include machine learning, object detection, localization \& tracking, action recognition. 
\end{IEEEbiography}

\begin{IEEEbiography}[{\includegraphics[width=1in,height=1.25in,clip,keepaspectratio]{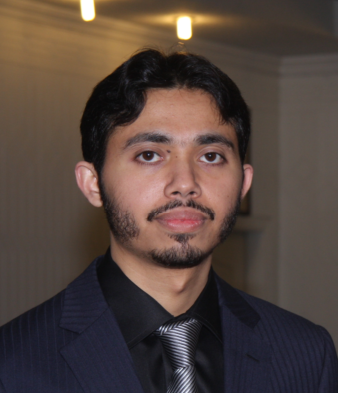}}]{Naveed Akhtar} 
received his PhD in Computer Vision from The University of Western Australia (UWA) and Master degree in Computer Science from Hochschule Bonn-Rhein-Sieg, Germany (HBRS). His research in  Computer Vision and Pattern Recognition has been published in  prestigious venues of the field, including IEEE CVPR and IEEE TPAMI. He has also served as a reviewer for these venues. During his PhD, he was  recipient of multiple scholarships, and  runner-up for the Canon Extreme Imaging Competition in 2015. 
Currently, he is a Research Fellow at UWA since July 2017. Previously, he has also served on the same position at the Australian National University for one year. His current research interests include people counting and tracking, adversarial machine learning, and hyperspectral image analysis.
\end{IEEEbiography}

\begin{IEEEbiography}[{\includegraphics[width=1in,height=1.25in,clip,keepaspectratio]{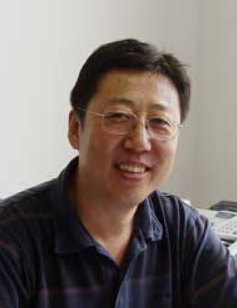}}]{HuanSheng Song} 
received the B.S. and M.S. degrees in communication and electronic systems and the Ph.D. degree in information and communication engineering from Xi’an Jiaotong University, Xi’an, China, in 1985, 1988, and 1996, respectively. Since 2004, he has been with the Information Engineering Institute, Chang’an University, Xi’an, where he became a Professor in 2006 and was nominated as the Dean in 2012. He has been involved in research on intelligent transportation systems for many years and has led a research team to develop a vehicle license plate reader and a traffic event detection system based on videos, which has brought about complete industrialization. His current research interests include image processing and recognition, as well as intelligent transportation systems.
\end{IEEEbiography}

\begin{IEEEbiography}[{\includegraphics[width=1in,height=1.25in,clip,keepaspectratio]{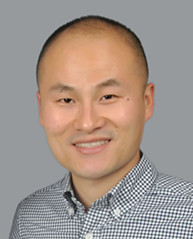}}]{ChaoYang Zhang} received his Ph.D. and B.S. degree in Applied Mathematics from Northwestern Polytechnical University (NWPU), Xi’an, China, in 2013 and 2007, respectively. He visited the Department of Computer Science Centre, at the University of Alberta, Edmonton, Canada, from 2011-2012. He is currently working as a teaching fellow at Chang’an University, Xi’an, China, from 2014. His current interests include linear and nonlinear feature extraction methods, pattern recognition, and remote sensing image processing.
\end{IEEEbiography}

\begin{IEEEbiography}[{\includegraphics[width=1in,height=1.25in,clip,keepaspectratio]{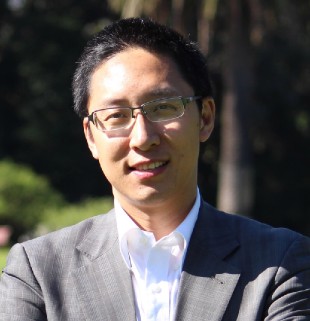}}]{JianXin Li}received received his PhD degree in computer science, from the Swinburne University of Technology, Australia, in 2009. He has been invited to be the PC-chair, proceedings chair, general chair in international conferences and workshops~-ADMA2016, WWW2017, DASFAA2018, PC member in SIGMOD2017, CIKM2018, ICDM2018, ICDE2019, and reviewer in top journals like TKDE. His research interests include database query processing \& optimization, social network analytics, and traffic network data processing.
\end{IEEEbiography}

\begin{IEEEbiography}[{\includegraphics[width=1in,height=1.25in,clip,keepaspectratio]{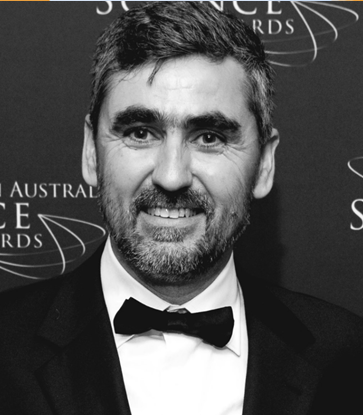}}]{Ajmal Mian} completed his PhD from The University of Western Australia in 2006 with distinction and received the Australasian Distinguished Doctoral Dissertation Award from Computing Research and Education Association of Australasia. He received the prestigious Australian Postdoctoral and Australian Research Fellowships in 2008 and 2011 respectively. He received the UWA Outstanding Young Investigator Award in 2011, the West Australian Early Career Scientist of the Year award in 2012 and the Vice-Chancellors Mid-Career Research Award in 2014. He has secured seven Australian Research Council grants, a National Health and Medical Research Council grant and a DAAD German Australian Cooperation grant. He is currently an Associate Professor of Computer Science at The University of Western Australia. His research interests include computer vision, machine learning, face recognition, 3D shape analysis and hyperspectral image analysis.
\end{IEEEbiography}

\end{document}